%% file: root.tex
\pdfoutput=1
\documentclass[letterpaper, 10 pt, conference]{ieeeconf}  

\IEEEoverridecommandlockouts                              

\input{includes/packages}
\input{includes/macros}

\title{\LARGE \bf
FIReStereo: Forest InfraRed Stereo Dataset for UAS \\
Depth Perception in Visually Degraded Environments
}

\author{Devansh Dhrafani$^{1*}$, Yifei Liu$^{2*}$, Andrew Jong$^{2}$, Ukcheol Shin$^{2}$, \\
 Yao He$^{2}$, Tyler Harp$^{2}$, Yaoyu Hu$^{2}$, Jean Oh$^{2}$, Sebastian Scherer$^{2}$
\thanks{* Denotes equal contribution}
\thanks{$^{1}$ Devansh Dhrafani is with Carnegie Mellon University, College of
Engineering, Department of Mechanical Engineering, Pittsburgh, PA
        {\tt\small \href{mailto:ddhrafan@andrew.cmu.edu}{ddhrafan@andrew.cmu.edu}}}%
\thanks{$^{2}$ Yifei Liu, Andrew Jong, Ukcheol Shin, Yao He, Tyler Harp, Yaoyu Hu, Jean Oh, Sebastian Scherer 
are with Carnegie Mellon University, School of Computer Science, the Robotics Institute, Pittsburgh, PA
        {\tt\small (\href{mailto:yifeil5@andrew.cmu.edu}{yifeil5}, 
    \href{mailto:ajong@andrew.cmu.edu}{ajong},
    \href{mailto:ushin@andrew.cmu.edu}{ushin}, 
    \href{mailto:yaohe@andrew.cmu.edu}{yaohe}, 
    \href{mailto:tharp@andrew.cmu.edu}{tharp},
    \href{mailto:yaoyuh@andrew.cmu.edu}{yaoyuh}, 
    \href{mailto:hyaejino@andrew.cmu.edu}{hyaejino}, 
    \href{mailto:basti@andrew.cmu.edu}{basti}\}@andrew.cmu.edu)}%
}}

\begin{document}

\maketitle
\pagestyle{empty}

\input{text/00-abstract}

\input{text/01-intro}

\input{text/02-related}

\input{text/03-dataset}

\input{text/04-evaluation}

\input{text/05-results}

\input{text/06-discussion}

\input{text/07-conclusion}

\input{text/08-acknowledgement}





\bibliographystyle{IEEEtran}
\bibliography{IEEEabrv, root} 


\end{document}

%% file: includes/packages.tex
\usepackage{graphicx}
\usepackage[dvipsnames]{xcolor, colortbl}
\usepackage{times} 
\usepackage{tikz}
\usepackage{amsmath} 
\usepackage{bm}  
\usepackage{amssymb}  
\usepackage{arydshln} 
\usepackage{mathrsfs}
\usepackage{eucal}
\usepackage{capt-of}
\usepackage[accsupp]{axessibility}  
\usepackage{float}
\usepackage{pifont}
\usepackage{caption}
\usepackage{subcaption}
\usepackage[breaklinks,colorlinks,bookmarks]{hyperref}
\usepackage{booktabs,arydshln}
\makeatletter
\def\adl@drawiv#1#2#3{%
        \hskip.5\tabcolsep
        \xleaders#3{#2.5\@tempdimb #1{1}#2.5\@tempdimb}%
                #2\z@ plus1fil minus1fil\relax
        \hskip.5\tabcolsep}
\newcommand{\cdashlinelr}[1]{%
  \noalign{\vskip\aboverulesep
           \global\let\@dashdrawstore\adl@draw
           \global\let\adl@draw\adl@drawiv}
  \cdashline{#1}
  \noalign{\global\let\adl@draw\@dashdrawstore
           \vskip\belowrulesep}}
\makeatother
\usepackage{boldline}  
\usepackage{multirow}
\usepackage{booktabs}
\usepackage{tabularray}
\usepackage{adjustbox}
\usepackage{siunitx}
\usepackage{hhline}
\usepackage{cite}

%% file: includes/macros.tex
\usepackage[capitalize]{cleveref}
\crefname{section}{Section}{Sections}
\crefname{table}{Table}{Tables}



\newcommand*{\subfigref}[2][]{%
  Fig. \hyperref[{fig:#2}]{%
    \ref*{fig:#2}%
    \ifx\\#1\\%
    \else
      \,#1%
    \fi
  }%
}

%% file: text/00-abstract.tex
\begin{abstract}
\label{abstract}
Robust depth perception in visually-degraded environments is crucial for autonomous aerial systems. Thermal imaging cameras, which capture infrared radiation, are robust to visual degradation. However, due to lack of a large-scale dataset, the use of thermal cameras for unmanned aerial system (UAS) depth perception has remained largely unexplored. This paper presents a stereo thermal depth perception dataset for autonomous aerial perception applications. The dataset consists of stereo thermal images, LiDAR, IMU and ground truth depth maps captured in urban and forest settings under diverse conditions like day, night, rain, and smoke. We benchmark representative stereo depth estimation algorithms, offering insights into their performance in degraded conditions. Models trained on our dataset generalize well to unseen smoky conditions, highlighting the robustness of stereo thermal imaging for depth perception. We aim for this work to enhance robotic perception in disaster scenarios, allowing for exploration and operations in previously unreachable areas. The dataset and source code are available at \url{https://firestereo.github.io}.


\end{abstract}

%% file: text/01-intro.tex
\section{INTRODUCTION}
\label{sec:intro}

Robotics has great potential to help in environments characterized by one or more of the five ``D's": dirty, dull, dangerous, difficult, and dear.
Examples of such environments include disaster response \cite{kanand2020wildfire}, construction \cite{de2017applicabilitycontruction}, mining \cite{padro2019monitoringmining}, and waste management \cite{sliusar2022dronewastemanagement}, which often contain dense and cluttered obstacles.
These environments may also suffer from visual degradations like smoke, dust, and darkness.
To operate effectively, robots need a robust and accurate geometric understanding of the scene, allowing them to estimate obstacle distance and localize within a perceived map.
However, these visual degradations interfere with common sensors used for perception: particulates affect LiDAR (smoke, dust, rain, snow), while darkness hinders RGB cameras.

One type of sensor that is well suited for perception in smoke, dust, and darkness is the long-wave infrared (LWIR) thermal camera.
Its infrared signal, in the 8 to 14 $\mu m$ range, bypasses smoke and dust as shown in \cref{fig:degradation}, and the heat signature of the environment still radiates in visual darkness. 

\input{figures/summary}

However, the development of thermal perception for cluttered environments is stalled by a scarcity of relevant datasets. Currently, the available thermal datasets are limited: some lack stereo camera pairs \cite{lee2022vivid++, dai2021multispectral, li2022odombeyondvision} that are necessary for geometric understanding, while others are restricted to urban road scenes \cite{shin2023ms2, yun2022sthereo}.

We present a new dataset, \emph{FIReStereo (Forest InfraRed Stereo Dataset)} for small UAS, to assist with developing depth estimation algorithms for sensors suited for visually degraded environments. 
This dataset was particularly motivated by smoky wildfires in the Wildland Urban Interface (WUI). The wildland urban interface describes where wilderness meets the urban population, and is where the most risk to human life and property damage tends to occur. 
Small Unmanned Aerial Systems (sUAS) have great potential to provide fine-grained situational awareness in such chaotic environments. By flying down into the smoke to assess the scene, an sUAS can help search for vulnerable people and triage the response.

\input{figures/degradation}

As such, our dataset \emph{FIReStereo} was collected through trajectories of cluttered environments that mimic the wildland urban interface. 
The trajectories were collected at a height below the tree canopy to simulate the perspective of a sUAS flying low to inspect the scene.
It consists of trees mixed with urban structures over a spectrum of visual conditions, such as daylight, darkness, rain, and smoke released by smoke bombs.
To estimate depth in smoke, the dataset's input imagery is captured from two stereo thermal Teledyne FLIR Boson™ cameras. These are placed at a relatively small 24cm camera separation. This smaller camera separation better represents the capacity of sUAS and robots that may navigate the narrow passages of cluttered environments.

For the smokeless environments, we provide ground-truth metric depth collected from LiDAR, that we densify using the Faster-LIO \cite{9718203fasterlio} algorithm.
The dataset contains the scene reconstruction and the estimated LiDAR-Inertial odometry of the sensor payload for state estimation.
However, since weight and size of LiDAR makes it impractical for most platforms constrained by size, weight, and power, we expect this dataset to also assist in advancement of thermal-inertial odometry and SLAM algorithms.

In summary, our contributions are as follows:
\begin{itemize}
    \item We contribute a novel dataset of visually-degraded environments to aid with metric depth estimation; in particular, with stereo infrared cameras in wildland-urban interface environments consisting of trees and urban structures.
    \item Our dataset contributes four distinct environment types (urban, mixed, and wilderness) with varying weather conditions (day, night, rain, cloud cover, smoke, fire).
    \item All sensor data is synchronized: rectified thermal stereo, LiDAR, and IMU information.
    \item We use our dataset to train representative supervised stereo depth estimation models, and show generalization to performance in visually degraded environments not present in the training data.
\end{itemize}

%% file: figures/summary.tex
\begin{figure}[ht]
    \centering
    \includegraphics[width=0.5\textwidth]{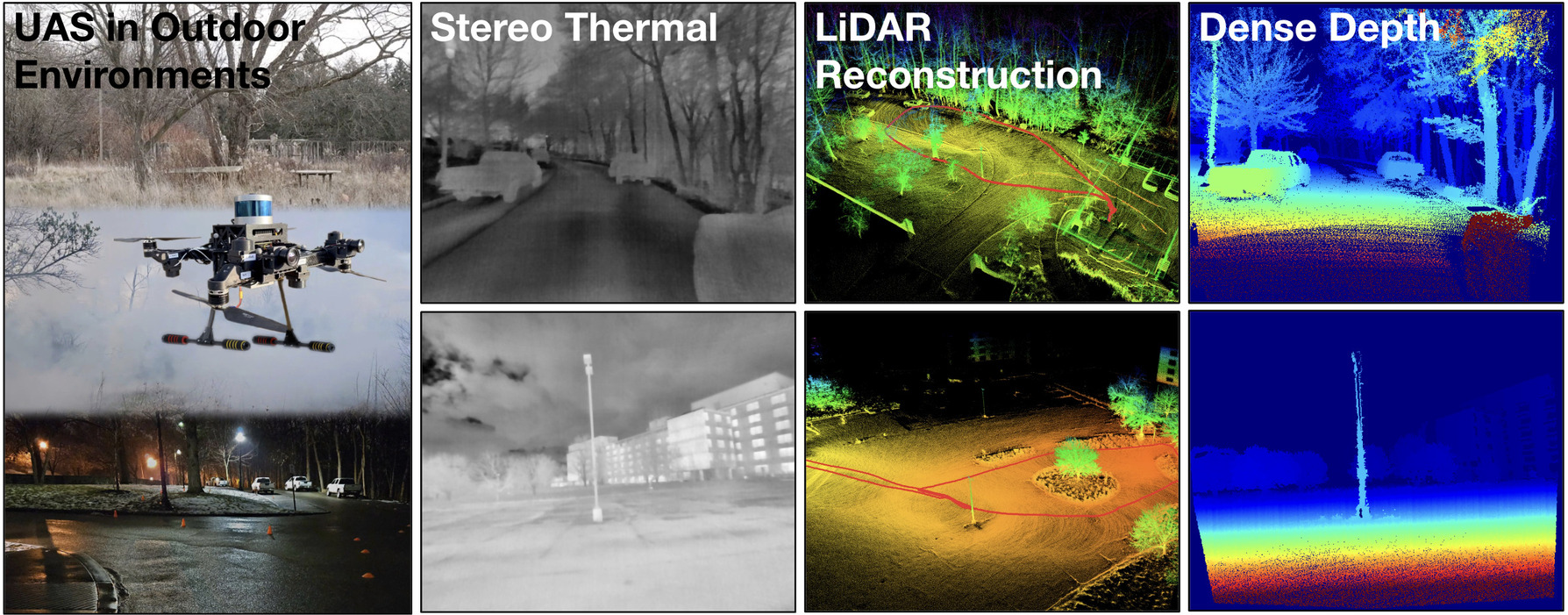}
    \caption{Overview of the UAS collection setup, various outdoor environments, and sensor modalities in the \textit{FIReStereo} Dataset.}
    \vspace{-4pt}
    \label{fig:summary}
\end{figure}

%% file: figures/degradation.tex
\begin{figure}[htbp]
\centering
\renewcommand\arraystretch{0} 
  \setlength{\tabcolsep}{1pt} 
  \begin{tabular}{ccc}
    Visual &
    LiDAR &
    Thermal \\
    \includegraphics[width=0.33\linewidth]{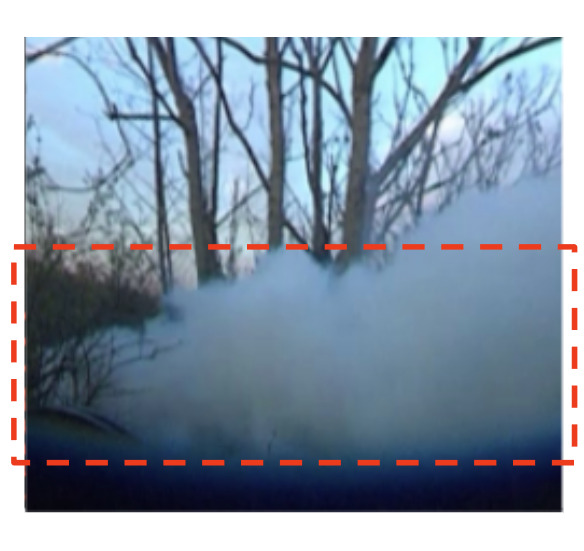} &
    \includegraphics[width=0.33\linewidth]{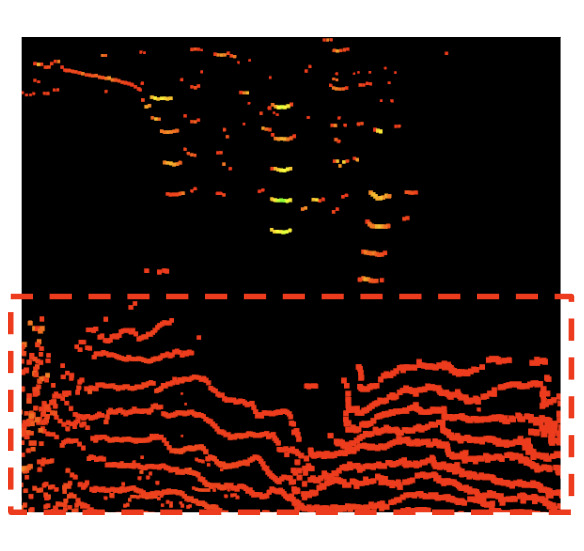} &
    \includegraphics[width=0.33\linewidth]{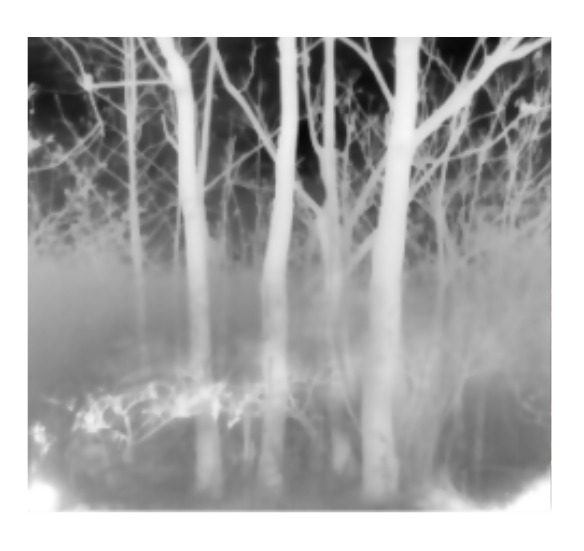} \\
  \end{tabular}

\caption{Comparison of sensors under smoke. Visual and LiDAR sensors are severely affected by particulates, while long-wave infrared is able to bypass the infrared signature from smoke, yielding clear visibility of the environment.   
} 
\vspace{-6pt}
\label{fig:degradation}
\end{figure}

%% file: text/02-related.tex
\section{RELATED WORK}
\label{sec:related}

\subsection{Metric depth estimation}

Real-time inference of metric distance to obstacles is required for online unmanned aerial system (UAS) navigation.
A representative method to estimate metric-scale depth maps is to convert a stereo image's disparity map into depth maps using the camera focal length $f$ and baseline $B$ (i.e. seperation between cameras). The disparity map encodes depth information of a scene by measuring the pixel displacement between corresponding points in stereo images. The conversion from disparity $d$ to depth $Z$ is given by:

\begin{align}
    Z = \frac{f \cdot B}{d}
    \label{eq:disp2depth}
\end{align}

Recent stereo-matching methods achieve high accuracy and efficiency using neural architectures to estimate disparity, categorized into 3D \cite{mayer2016large, liang2018learning, xu2020aanet} and 4D cost volume methods~\cite{chang2018pyramidpsmnet,guo2019groupgwcnet,xu2022attention}. While 3D methods are more efficient by constructing single-channel cost volumes, 4D methods show more accurate disparity maps by constructing multiple-channel cost volumes and aggregating them via 3D convolutions. However, both approaches face a high computational burden, limiting their suitability for real-world applications and on-device inference.
Therefore, another research area is designing a lightweight stereo-matching network for on-device inference \cite{shamsafar2022mobilestereonet, xu2023accuratefastacv}. Models like MobileStereoNet \cite{shamsafar2022mobilestereonet} and FastACVNet \cite{xu2023accuratefastacv} achieve efficiency by leveraging lightweight architectures and novel cost volume formulations, reducing computational complexity. 

We train representative models from the discussed categories on our dataset to benchmark their performance and analyze strengths and limitations for real-time metric depth estimation on UAS platforms.

\input{tables/dataset_comp}

\subsection{Visual datasets}
Visual sensors capture the visible spectrum of light and are commonly used in robotics applications. A diverse large-scale image dataset paired with ground truth depths are crucial for training deep learning models to predict distance. Prominent datasets like KITTI \cite{kitti2013}, Oxford \cite{maddern20171oxford}, DDAD \cite{guizilini20203ddad}, Cityscape \cite{cordts2016cityscapes}, and nuScenes \cite{caesar2020nuscenes} provide such pairs but are limited to driving scenes with constrained motion patterns, typically in the forward direction with small left or right turns \cite{wang2020tartanair}. Training on these datasets often lead to models that struggle to generalize to UAS tasks, which involve more aggressive, multi-axial motion and smaller stereo camera baselines. 

InteriorNet \cite{li2018interiornet} and SceneNet \cite{mccormac2017scenenet} are synthetically generated whereas TUM RGB-D \cite{sturm2012tumrgbd} and ScanNet \cite{dai2017scannet} are real world indoor datasets with aggressive motion patterns and small camera separation. However, the latter three lack stereo image pairs, critical for metric depth estimation. Although InteriorNet provides stereo pairs, it is limited to indoor environemnts, leading to poor generalization in outdoor settings. TartanAir \cite{wang2020tartanair} is a synthetic UAS dataset that provides stereo images with ground truth depth across various indoor and outdoor urban, rural, and nature scenes, resembling potential UAS diaster response scenarios. However, visible-spectrum cameras remain vulnerable to factors like illumination, dust, and smoke. Thermal cameras, being more robust to such conditions, have gained increasing attention in the robotics community. 

\subsection{Thermal datasets}

Thermal cameras, which capture infrared radiation from object surfaces, have gained popularity because of their robustness to adverse conditions, widespread availability and unique modality information (i.e., temperature). As thermal cameras became more accessible, many thermal datasets have been proposed. MultiSpectralMotion \cite{dai2021multispectral} is an indoor-outdoor handheld dataset with thermal image and ground truth depth map. OdomBeyondVision \cite{li2022odombeyondvision} is an indoor-only thermal dataset with handheld, unmanned ground vehicle (UGV) and UAS movements. ViViD++ \cite{lee2022vivid++} is an outdoor thermal dataset with vehicle and handheld movements. SubT-MRS \cite{zhao2023subtmrs} is an outdoor dataset with vehicle, UAS, legged and handheld motion in varying levels of visual degradation. SubT-MRS is the closest to our environmental needs but it does not contain wildland forest images. Moreover, all the above datasets include single thermal camera views and so cannot be used to train stereo depth estimation models. 

STheReO \cite{yun2022sthereo} and MS2 \cite{shin2023ms2} are recent datasets with outdoor stereo thermal images. MS2 provides images in a wide variety of environmental conditions with visually-degraded scenes like rain. But both of these datasets are limited to urban driving scenes and suffer from the same problems as visual driving datasets -- limited motion patterns and large camera baseline. In addition, all above datasets are restricted to urban scenes, and do not cover wild outdoor environments like dense forests.

A comprehensive comparison between datasets is shown in Table \ref{tab:dataset_comparison}. As discussed, existing datasets are insufficient to investigate environmentally robust metric-scale depth estimation for sUAS under visually-degraded environments. This motivates our dataset.



%% file: tables/dataset_comp.tex
\newcommand{\cmark}{\ding{51}}
\newcommand{\xmark}{\ding{53}}

\begin{table*}[t]
\centering
\vspace{8pt}
\caption{A comparison with previous alternative thermal datasets. Our proposed dataset captures aerial stereo thermal images in urban/forest environments under smoke and rainy conditions.}
\label{tab:dataset_comparison}
\resizebox{\linewidth}{!}{%
\begin{tabular}{|lccccccccccc|} 
\hline
\multicolumn{1}{|c|}{\multirow{2}{*}{Dataset}} & \multicolumn{1}{c|}{\multirow{2}{*}{Year}} & \multicolumn{1}{c|}{\multirow{2}{*}{Depth Sensor}} & \multicolumn{3}{c|}{Movement Pattern}                                                     & \multicolumn{2}{c|}{Thermal}                            & \multicolumn{2}{c|}{Environment}                         & \multicolumn{2}{c|}{Conditions}                   \\ 
\cline{4-12}
\multicolumn{1}{|c|}{}                         & \multicolumn{1}{c|}{}                      & \multicolumn{1}{c|}{}                              & \multicolumn{1}{c|}{Handheld} & \multicolumn{1}{c|}{Ground} & \multicolumn{1}{c|}{Aerial} & \multicolumn{1}{c|}{Mono} & \multicolumn{1}{c|}{Stereo} & \multicolumn{1}{c|}{Urban} & \multicolumn{1}{c|}{Forest} & \multicolumn{1}{c|}{Smoke} & Rain                 \\ 
\hline
ViViD++ \cite{lee2022vivid++}                  & 2022                                       & LiDAR                                              & \cmark           & \cmark         & \xmark                   & \cmark       & \xmark                   & \cmark        & \xmark                   & \xmark                  & \xmark            \\
MultiSpectralMotion \cite{dai2021multispectral} & 2021                                       & RGB-D                                              & \cmark           & \xmark                   & \xmark                   & \cmark       & \xmark                   & \cmark        & \xmark                   & \xmark                  & \xmark            \\
OdomBeyondVision \cite{li2022odombeyondvision}  & 2022                                       & LiDAR                                              & \cmark           & \cmark         & \cmark         & \cmark       & \xmark                   & \xmark                  & \xmark                   & \xmark                  & \xmark            \\
STheReO \cite{yun2022sthereo}                 & 2022                                       & LiDAR                                              & \xmark                     & \cmark         & \xmark                   & \cmark       & \cmark         & \cmark        & \xmark                   & \xmark                  & \xmark            \\
MS2 \cite{shin2023ms2}                        & 2023                                       & LiDAR                                              & \xmark                     & \cmark         & \xmark                   & \cmark       & \cmark         & \cmark        & \xmark                   & \xmark                  & \cmark  \\
SubT-MRS \cite{zhao2023subtmrs}               & 2023                                       & RGB-D, LiDAR                                       & \cmark           & \cmark         & \cmark         & \cmark       & \xmark                   & \cmark        & \xmark                   & \cmark        & \cmark  \\ 
\hline
Ours                                           & 2024                                       & LiDAR                                              & \cmark           & \xmark                   & \cmark         & \cmark       & \cmark         & \cmark        & \cmark         & \cmark        & \cmark  \\
\hline
\end{tabular}
}
\end{table*}

%% file: text/03-dataset.tex
\section{FIRESTEREO DATASET}

\subsection{System overview} 

\input{figures/dataset/system/data_collection_system}

\textbf{Hardware setup} 
We designed a data collection platform comprising a pair of stereo thermal cameras, a LiDAR, and an inertial measurement unit (IMU), as illustrated in ~\cref{fig:data_collection_setup}. These sensors are mounted on an unmanned aerial vehicle (UAV) platform, which facilitates data collection during handheld experiments and UAV flights. Sensor specifications are presented in ~\cref{tab:sensors}. The stereo thermal pair is oriented in a forward-facing direction with a 24.6 cm baseline, the LiDAR is mounted on top of the UAV, and an onboard NVIDIA® Jetson AGX Orin™ computer is connected to the sensors.

\input{tables/table-sensor_descriptions}

\textbf{Time Synchronization} Accurate time synchronization is a critical prerequisite for various tasks that involve multiple sensors. Our synchronization of the sensors is carried out using the pulse-per-second (PPS) technique. The IMU, LiDAR, and thermal cameras are synchronized by electronic pulses from the CPU clock.

\subsection{Data collection} 
Sensor data was recorded through the Robot Operating System (ROS) framework running onboard the Orin. In addition to the processed dataset, we provide the corresponding dataloader, calibration files, and rosbags covering all data collection sessions. For Camera-IMU calibration, we use Kalibr \cite{furgale2012continuouskalibr} with a radial-tangential distortion model and heated checkerboard target \cite{shivakumar2020pst900thermalcalib}. For the FLIR Boson thermal camera, we capture the raw 16-bit data. The Flat Field Correction (FFC) of the thermal camera was set to manual and triggered at the start of each sequence, but not during collection. The data includes recordings from 4 distinct locations and 16 unique trajectories under various environmental conditions, including day, night, rain, cloud cover, and smoke. Smoke was emitted from training-grade smoke pots. All collected sensor data is time-synchronized with the CPU clock, making it also suitable to use for mapping and localization. A closed-loop trajectory was followed with the same initial and final position, making the data set suitable for testing loop closure and accumulated drift for mapping and localization. Examples of such trajectories may be found in the 3D reconstruction maps in \cref{fig:slam_maps}.

\input{figures/dataset/slam_maps/slam_maps}

\subsection{Ground truth depth/disparity generation}
Supervised training of depth perception models requires Ground-Truth (GT) depth or disparity map outputs that correspond to each pair of input stereo images. 
Conventionally, LiDAR-based datasets \cite{kitti2013, shin2023ms2} produce these maps by aggregating multiple LiDAR scans via the Iterative Closest Point (ICP) algorithm \cite{besl1992methodicp}. 
Although this methodology is effective for capturing large obstacles such as vehicles in driving scenarios, it results in sparsely populated depth maps. This sparsity becomes particularly evident when the dataset includes finer obstacles like tree branches and thin trunks, which are imperative to detect for sUAS navigation. To address this limitation and enhance the density of the depth map, we employ Simultaneous Localization and Mapping (SLAM), which enables the generation of more comprehensive depth maps capable of accurately representing thin obstacles such as branches, poles, and tree trunks. The ground truth depth maps can be converted to disparity maps using Eq \ref{eq:disp2depth}.

\input{figures/dataset/environment_photos/environment_photos}

We use Faster-LIO \cite{9718203fasterlio} to estimate the UAS trajectory and generate a dense point cloud of the environment. Since the thermal images are obtained at a higher frequency than LiDAR, we interpolate the odometry robot pose to match the thermal timestamps. We employ linear interpolation for positions and spherical linear interpolation \cite{shoemake1985animatingslerp} for orientation. Camera poses are found by transforming the odometry pose to camera frame using IMU-camera extrinsic. 

For each camera pose, the point cloud is projected to image frame with Open3D \cite{Zhou2018open3d}. Due to the sparse nature of the global point cloud map when projected into the camera frame, some points that should normally be occluded can become visible. To address this issue, we developed a 2D grid blocking technique in the image space. This method involves overlaying a 2x2 grid across the image with a stride of 1. Within each grid cell, we identify the point closest to the camera and set this as the minimum distance for all points in that cell. To further filter the points, we apply a left-right consistency check \cite{shin2023ms2}, reconstructing each image by warping the other using the disparity map and excluding pixels with significant reconstruction errors.

Due to the interference of smoke with LiDAR functionality, resulting in noisy depth measurements, ground-truth depth maps were not produced for smoky conditions. However, we have included the LiDAR data for these conditions within the rosbags.

\subsection{Data description}
The four collection sites differ significantly in tree density, vegetation, and proximity to urban structures such as parking lots and vehicles. We analyze the characteristics of our dataset quantitatively and qualitatively. 

\textbf{Quantitative Data Description} 
The processed \emph{FIReStereo} dataset contains 204,594 stereo thermal images total across all environments. 
29\% are in urban environment, 15\% are in mixed environment, 56\% are in wilderness environment with dense trees.
84\% of the images were collected in day-time and the rest were during night-time. Obstacles were measured at a median depth of 7.40 m with quartiles q1 = 5.17 m and q3 = 10.52 m, which falls within the typical range for UAS obstacle avoidance.
42\% stereo thermal pairs are smokeless, while 58\% contain smoke.
Of the smokeless images, we label 35,706 with depth-map pairs annotated from LiDAR and Faster-LIO.

For the depth-annotated images, we train supervised models representative of the depth-estimation literature on the data containing LiDAR ground truth. For model training and evaluation, we divide the depth-annotated data into train,
validation, and test subsets, with approximately a 70\%, 15\%, 15\% split, resulting in 25,653, 5,032, and 5,021 images, respectively. 
We ensured that each subset has a similar distribution of weather conditions and the type of scene. As some sequences were significantly longer than others, we split the sequences into two or more parts and then distributed the parts to one of the three subsets. 
\cref{tab:firestereo_subsets} describes the number of images per run, weather condition, and environment description.

\input{tables/FIReStereo_subsets}

\input{figures/dataset/data_variety/data_variety}

\textbf{Qualitative Data Description}
~\cref{fig:env_photos} and ~\cref{fig:data_variety} illustrate examples of the variety of data collected.
The Hawkins sequences feature scenes of dense forests and urban structures, recorded during cloudy, windy daytime conditions. In the first two sequences, the UAS navigates around dense trees with varying branch thickness, many of which are bare due to the spring season, alongside some evergreens. The next two sequences were captured in urban environments, with sequence 3 showing a parking lot with sparse trees and includes views of a thin pole, trees, and buildings representing typical urban obstacles that a UAS might face in response to a wildfire disaster. Sequence 4 captures a car, pole and distant trees. Sequence 5 replicated a disaster scenario with an upside-down car engulfed in dense smoke. We later show inference on this data to demonstrate generalizability to unseen smoke-filled data.

The Frick experimental sequences were recorded during the night and under rainy conditions. The captured temperature range for these sequences is much lower and is evident from the darker thermal images. These sequences feature bare trees in varying sparsity, vehicles, poles, roads and buildings. 

The Gascola sequences were recorded in heavily degraded wilderness, featuring dense smoke, night-time, dense trees and bushes. These conditions were chosen to simulate the wildfire disaster response scenario in which the UAS must navigate through a cluttered forest environment with extreme visual degradation. We show that depth estimation models trained on smokeless data is able to generalize to these smoke-filled data.

Firesgl sequences were recorded during actual prescribed fires, capturing flames, embers, smoke, and dense trees. These sequences represent highly realistic wildfire scenarios, providing critical data for testing depth estimation in extreme environments.

%% file: figures/dataset/system/data_collection_system.tex
\begin{figure}[htbp]
\centering
\includegraphics[width=0.92\linewidth]{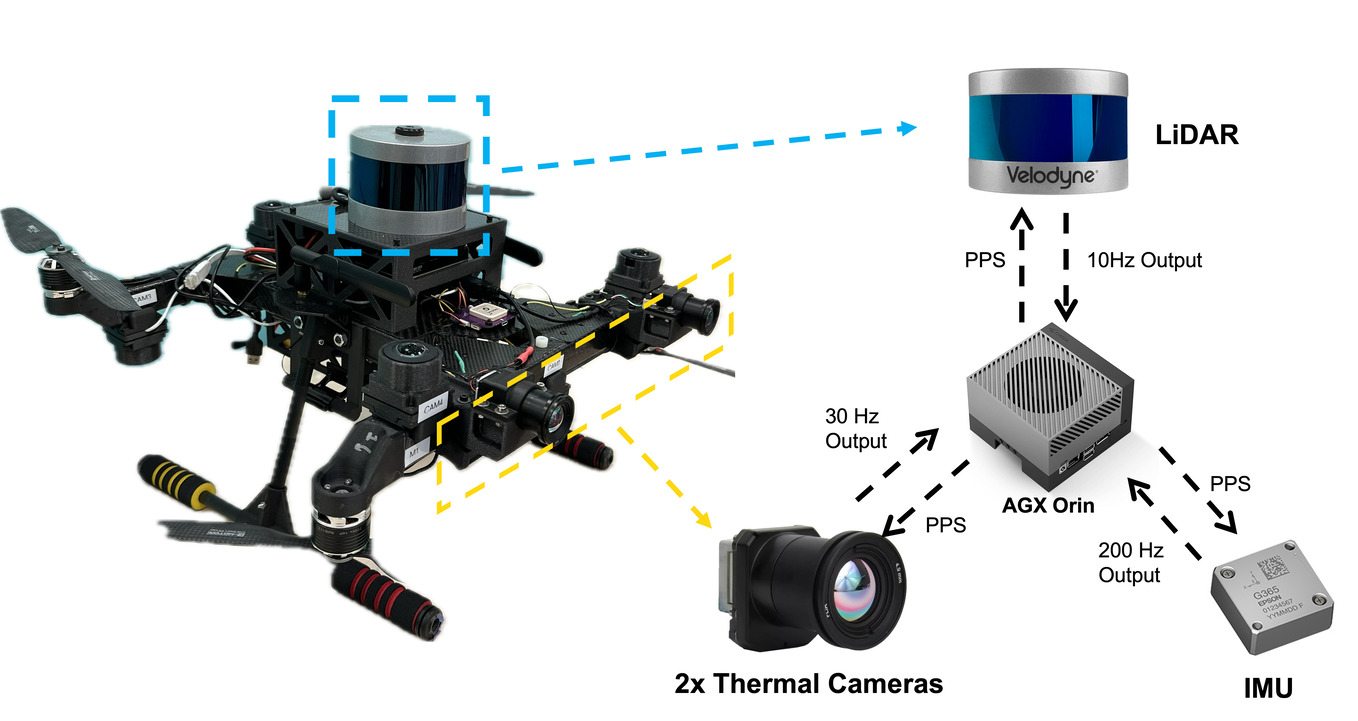}
\caption{Data collection setup with stereo thermal pair, LiDAR, IMU, and AGX Orin.}
\vspace{-10pt}
\label{fig:data_collection_setup}
\end{figure}

%% file: tables/table-sensor_descriptions.tex
\begin{table}[ht] 
\centering 
\caption{Sensor specification for the thermal stereo system}
\resizebox{\columnwidth}{!}{
\begin{tabular}{c|ccc}
\toprule
 \textbf{Sensors} & \textbf{Model} & \textbf{Frame Rate} & \textbf{Characteristics}   \\
 \cmidrule{1-1} \cmidrule(lr{0.75em}){2-4}
    \multirow{4}{*}{Thermal Camera} & \multirow{4}{*}{FLIR Boson+ 640} &  \multirow{4}{*}{Max 30 fps} & 640×512pixel \\ & & & 4.9MM 95\textdegree HFOV \\ & & & 12 µm pixel pitch VOx microbolometer \\ & & & 16-bit Raw data \\
    \cmidrule{1-1} \cmidrule(lr{0.75em}){2-4}
    \multirow{3}{*}{LiDAR} & \multirow{3}{*}{Velodyne VLP16} &  \multirow{3}{*}{10 fps} & Accuracy: $\pm 3cm$ \\ & & & Measurement range:100m 95\textdegree HFOV \\ & & & 360\textdegree(H), $\pm 15$\textdegree (V) FoV  \\
    \cmidrule{1-1} \cmidrule(lr{0.75em}){2-4}
    \multirow{3}{*}{IMU} & \multirow{3}{*}{Epson G365} &  \multirow{3}{*}{200 fps} & 6-DoF \\ & & & Acceleration \\ & & & Gyroscope  \\
\bottomrule
\end{tabular}
}
\label{tab:sensors}
\end{table}

%% file: figures/dataset/slam_maps/slam_maps.tex
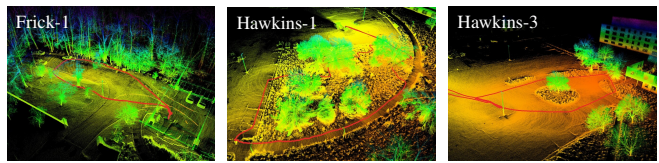
\begin{figure}[ht!]
\centering
\begin{minipage}{\columnwidth}
\begin{subfigure}{0.32\columnwidth}
    \begin{tikzpicture}
      \node[inner sep=0pt] (image) at (0,0) {\includegraphics[width=\linewidth]{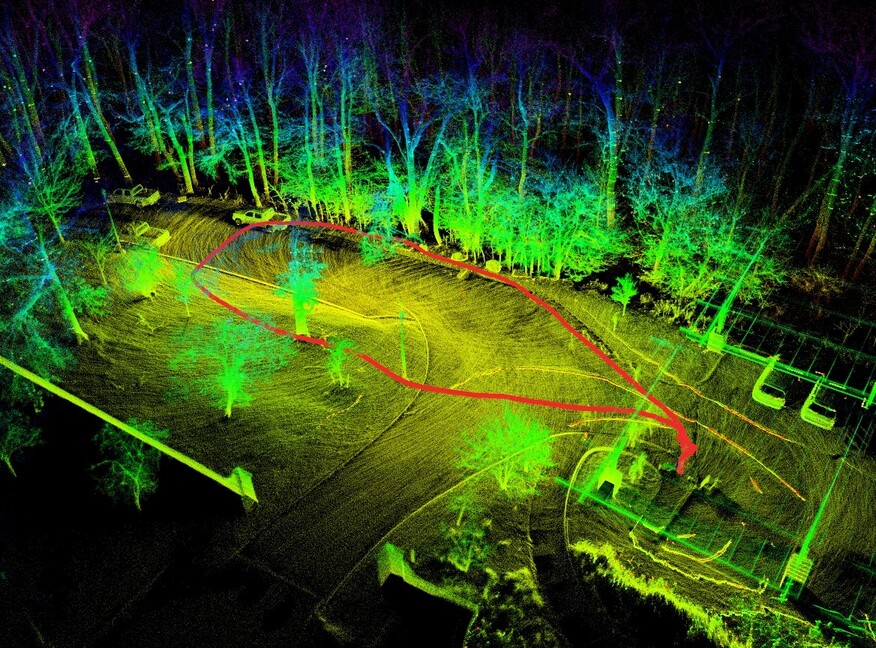}};
      \node[anchor=north west, white, font=\scriptsize] at (image.north west) {Frick-1};
    \end{tikzpicture}
\end{subfigure}
\hfill
\begin{subfigure}{0.32\columnwidth}
    \begin{tikzpicture}
      \node[inner sep=0pt] (image) at (0,0) {\includegraphics[width=\linewidth]{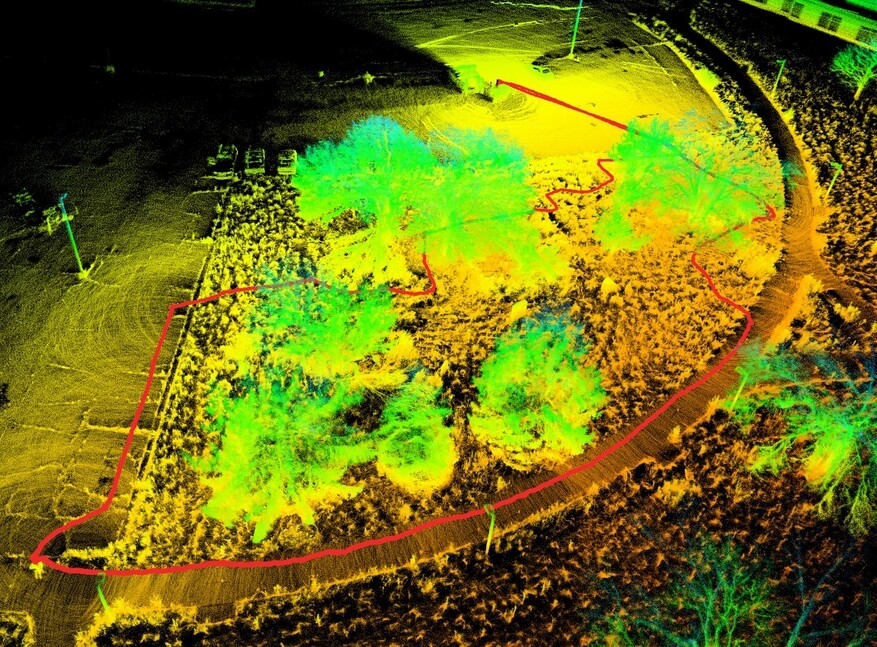}};
      \node[anchor=north west, white, font=\scriptsize] at (image.north west) {Hawkins-1};
    \end{tikzpicture}
\end{subfigure}
\hfill
\begin{subfigure}{0.32\columnwidth}
    \begin{tikzpicture}
      \node[inner sep=0pt] (image) at (0,0) {\includegraphics[width=\linewidth]{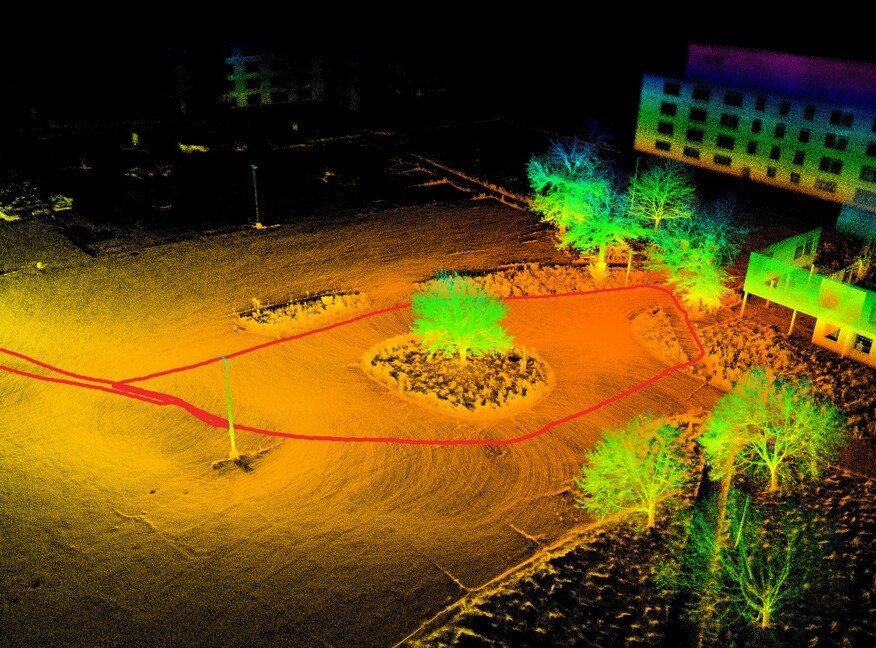}};
      \node[anchor=north west, white, font=\scriptsize] at (image.north west) {Hawkins-3};
    \end{tikzpicture}
\end{subfigure}
\end{minipage}
\caption{Maps generated using LiDAR-SLAM, which is used to obtain ground truth depth in \protect\cref{fig:data_variety}.}
\vspace{-10pt}
\label{fig:slam_maps}
\end{figure}

%% file: figures/dataset/environment_photos/environment_photos.tex

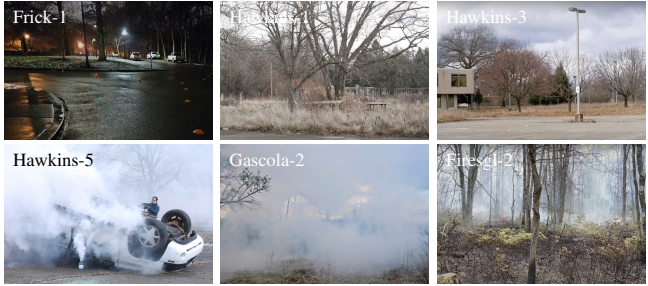
\begin{figure}[ht!]
\centering
\begin{minipage}{\columnwidth}
\begin{subfigure}{0.32\columnwidth}
    \begin{tikzpicture}
      \node[inner sep=0pt] (image) at (0,0) {\includegraphics[width=\linewidth]{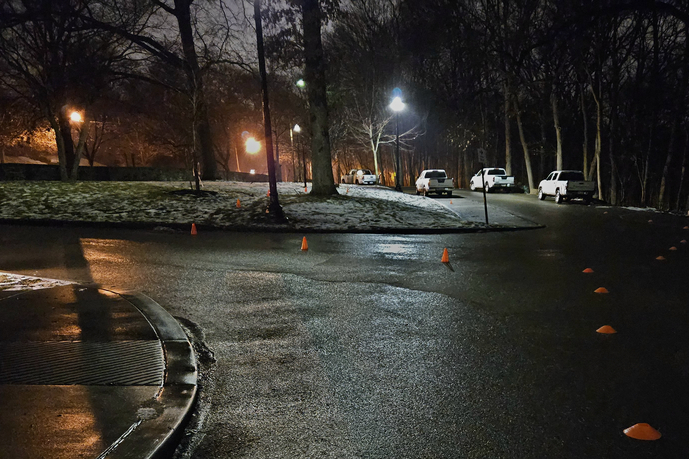}};
      \node[anchor=north west, white, font=\scriptsize] at (image.north west) {Frick-1};
    \end{tikzpicture}
\end{subfigure}
\hfill
\begin{subfigure}{0.32\columnwidth}
    \begin{tikzpicture}
      \node[inner sep=0pt] (image) at (0,0) {\includegraphics[width=\linewidth]{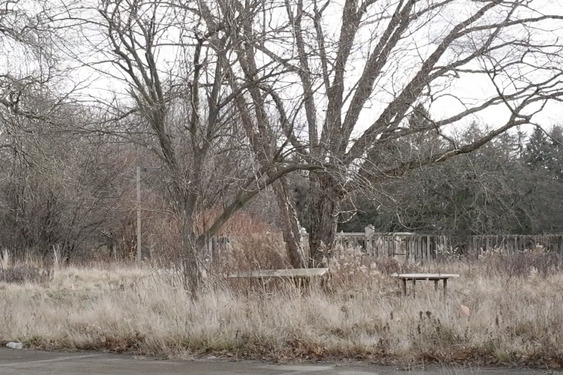}};
      \node[anchor=north west, white, font=\scriptsize] at (image.north west) {Hawkins-1};
    \end{tikzpicture}
\end{subfigure}
\hfill
\begin{subfigure}{0.32\columnwidth}
    \begin{tikzpicture}
      \node[inner sep=0pt] (image) at (0,0) {\includegraphics[width=\linewidth]{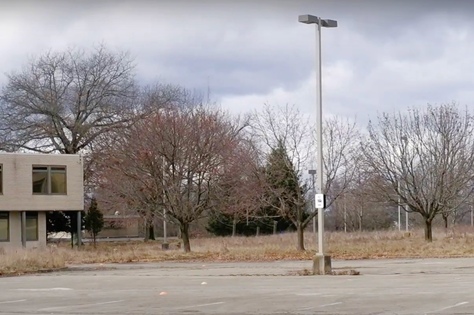}};
      \node[anchor=north west, white, font=\scriptsize] at (image.north west) {Hawkins-3};
    \end{tikzpicture}
\end{subfigure}
\hfill
\end{minipage}

\vspace{0.6mm}

\begin{minipage}{\columnwidth}
\begin{subfigure}{0.32\columnwidth}
    \begin{tikzpicture}
      \node[inner sep=0pt] (image) at (0,0) {\includegraphics[width=\linewidth]{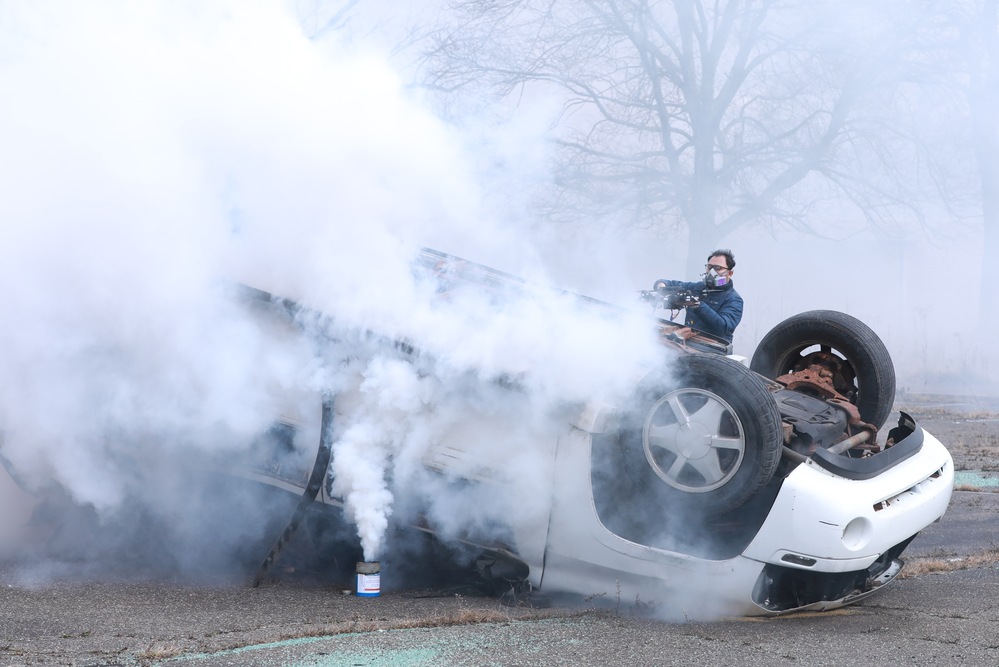}};
      \node[anchor=north west, black, font=\scriptsize] at (image.north west) {Hawkins-5};
    \end{tikzpicture}
\end{subfigure}
\hfill
\begin{subfigure}{0.32\columnwidth}
    \begin{tikzpicture}
      \node[inner sep=0pt] (image) at (0,0) {\includegraphics[width=\linewidth]{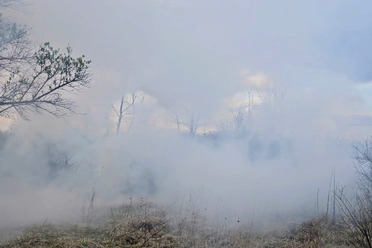}};
      \node[anchor=north west, white, font=\scriptsize] at (image.north west) {Gascola-2};
    \end{tikzpicture}
\end{subfigure}
\hfill
\begin{subfigure}{0.32\columnwidth}
    \begin{tikzpicture}
      \node[inner sep=0pt] (image) at (0,0) {\includegraphics[width=\linewidth]{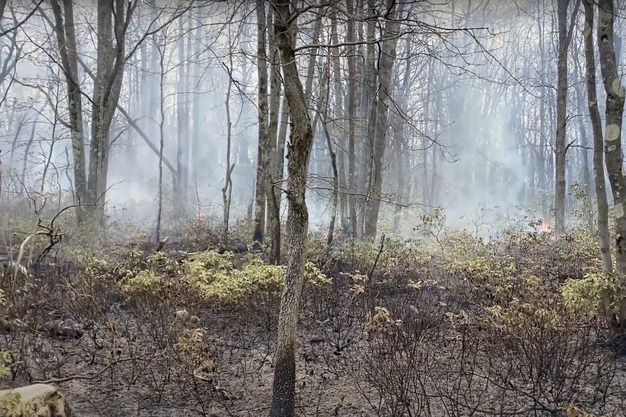}};
      \node[anchor=north west, white, font=\scriptsize] at (image.north west) {Firesgl-2};
    \end{tikzpicture}
\end{subfigure}
\hfill
\end{minipage}
\caption{Photos of different environments from our dataset featuring different tree densities, light conditions, and smoke conditions.}
\vspace{-6pt}
\label{fig:env_photos}
\end{figure}

%% file: tables/FIReStereo_subsets.tex
\begin{table}[!bh]
\centering
\caption{Comparison of FIReStereo data sequences}
\label{tab:firestereo_subsets}
\resizebox{\linewidth}{!}{%
\begin{tabular}{lrcccc} 
\toprule
\multirow{2}{*}{\textbf{Sequence}} & \multicolumn{1}{c}{\multirow{2}{*}{\textbf{\#Samples}}} & \multirow{2}{*}{\begin{tabular}[c]{@{}c@{}}\textbf{Depth}\\\textbf{Map}\end{tabular}} & \multicolumn{2}{c}{\textbf{Conditions}} & \multirow{2}{*}{\textbf{Environment}}  \\ 
\cline{4-5}& \multicolumn{1}{c}{}                                    &                                                                                       & \multicolumn{1}{l}{\textbf{Smoke}} & \multicolumn{1}{l}{\textbf{Rain}} &                                        \\ 
\hline
Hawkins-1 trees                    & 12,478                                                   & \cmark                                                                 &                                    &                                   & Mixed\\
Hawkins-2 trees                    & 10,180                                                   & \cmark                                                                 &                                    &                                   & Mixed\\
Hawkins-3 pole                     & 7,009                                                    & \cmark                                                                 &                                    &                                   & Mixed\\
Hawkins-4 upright car              & 4,127                                                    & \cmark                                                                 &                                    &                                   & Urban\\
Hawkins-5 flipped car smoke        & 25,713                                                   &                                                                                       & \cmark              &                                   & Urban\\
Hawkins-6 pole smoke               & 28,535                                                   &                                                                                       & \cmark              &                                   & Urban\\
Frick-1 night rain                 & 736& \cmark                                                                 &                                    & \cmark             & Mixed\\
Frick-2 night rain                 & 1,176                                                    & \cmark                                                                 &                                    & \cmark             & Mixed\\
Gascola-1 day                      & 8,617                                                    &                                                                                       &                                    &                                   & Dense Trees\\
Gascola-2 day+smoke                & 12,032                                                   &                                                                                       & \cmark              &                                   & Dense Trees\\
Gascola-3 static smoke             & 15,761                                                   &                                                                                       & \cmark              &                                   & Sparse Trees\\
Gascola-4 night                    & 11,908                                                   &                                                                                       &                                    &                                   & Dense Trees\\
Gascola-5 night smoke              & 12,307                                                   &                                                                                       & \cmark              &                                   & Dense Trees\\
 Firesgl-1-3 prescribed fire& 54,015& & \cmark& &Dense Trees\\
\bottomrule
\end{tabular}
}
\end{table}

%% file: figures/dataset/data_variety/data_variety.tex
\begin{figure*}[t]
  \vspace{12pt}
  \centering
  \renewcommand\arraystretch{0} 
  \setlength{\tabcolsep}{1pt} 
  \begin{tabular}{ccccccc}
    \includegraphics[width=0.14\linewidth]{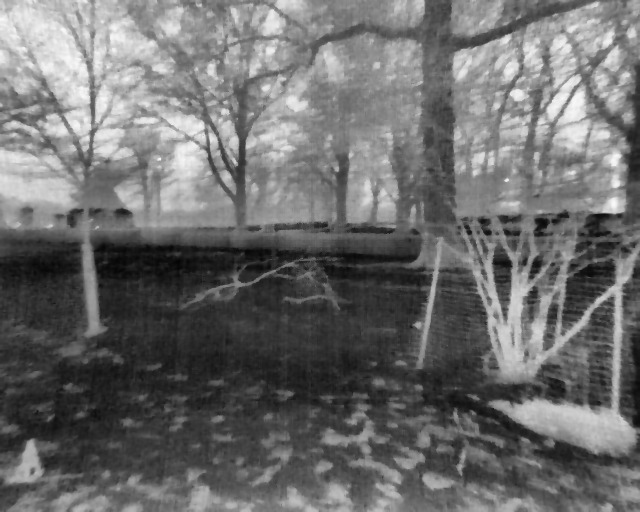} &
    \includegraphics[width=0.14\linewidth]{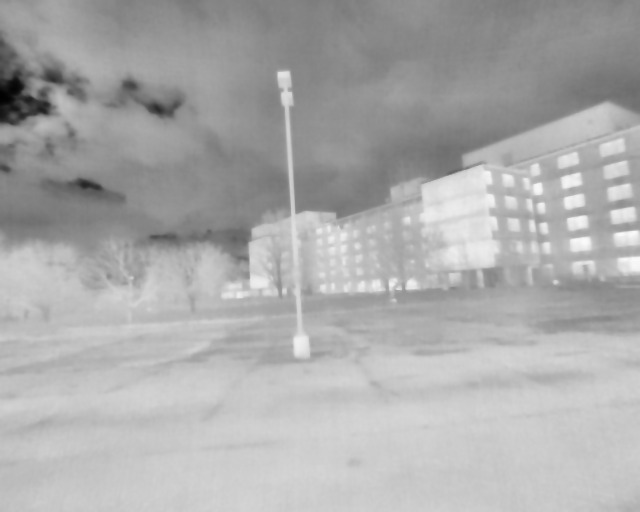} &
    \includegraphics[width=0.14\linewidth]{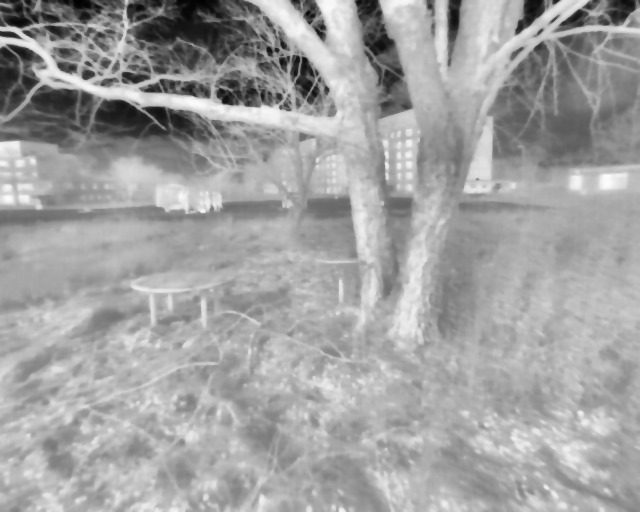} &
    \includegraphics[width=0.14\linewidth]{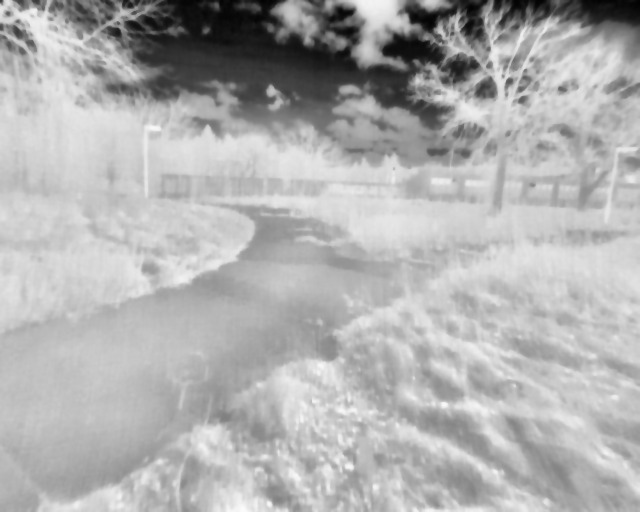} &
    \includegraphics[width=0.14\linewidth]{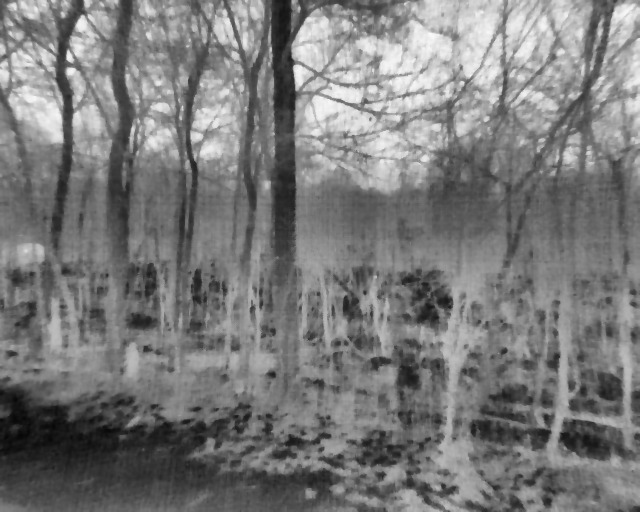} &
    \includegraphics[width=0.14\linewidth]{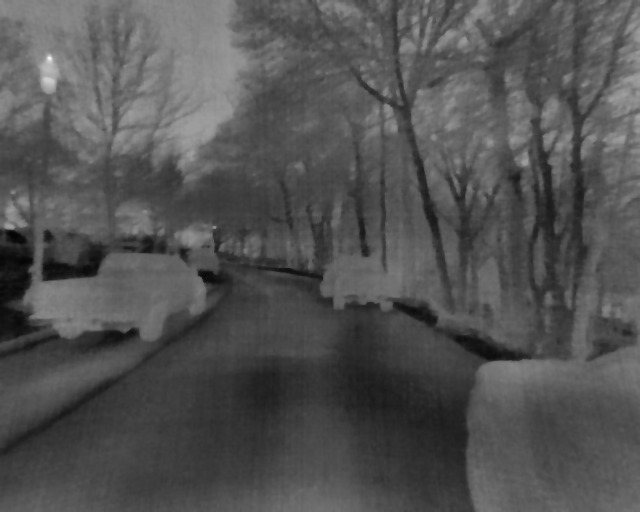} &
    \includegraphics[width=0.14\linewidth]{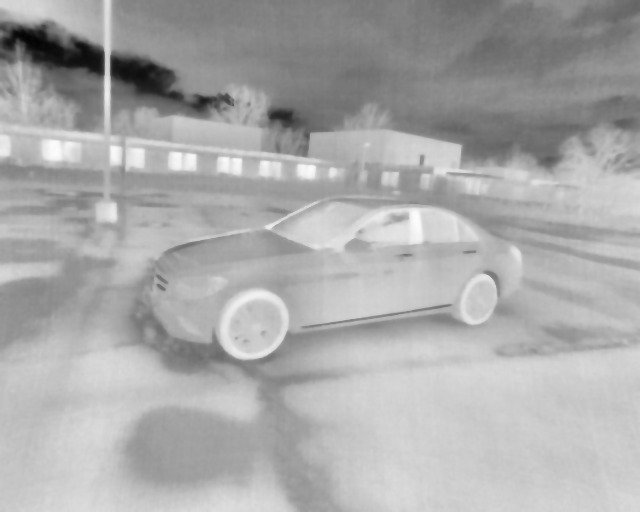} \\
    [2pt]
    \includegraphics[width=0.14\linewidth]{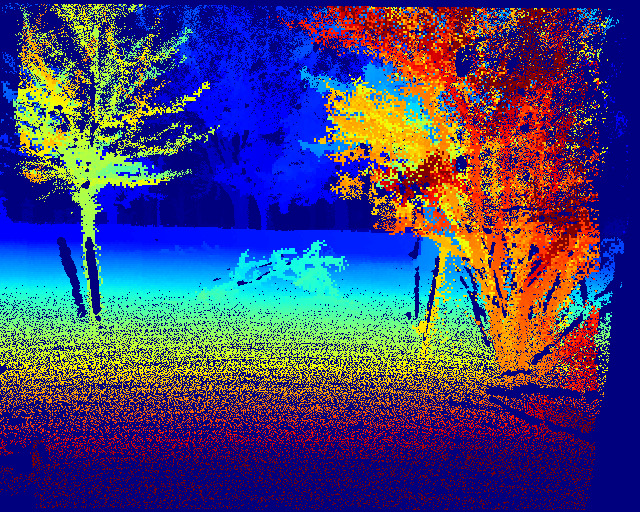} &
    \includegraphics[width=0.14\linewidth]{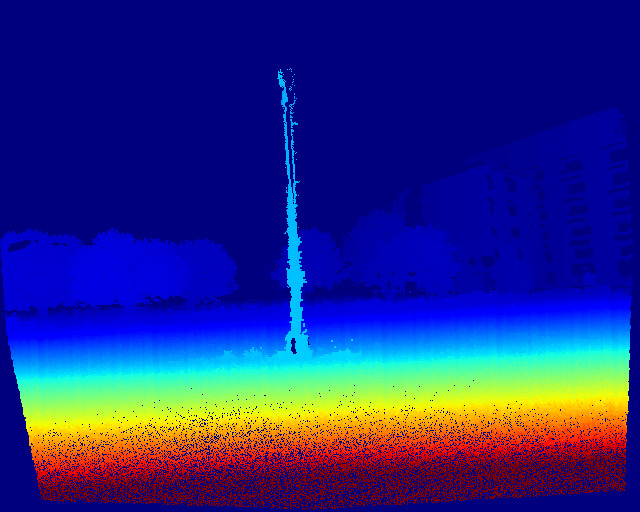} &
    \includegraphics[width=0.14\linewidth]{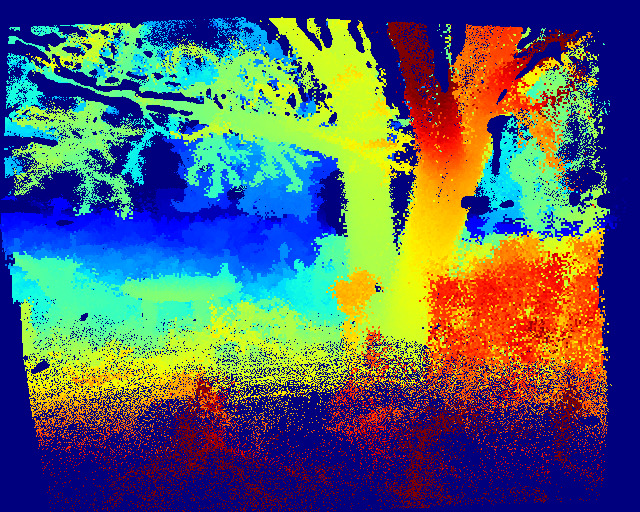} &
    \includegraphics[width=0.14\linewidth]{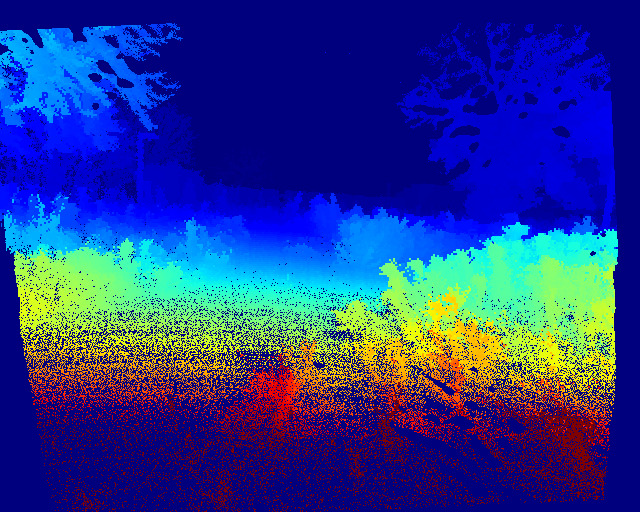} &
    \includegraphics[width=0.14\linewidth]{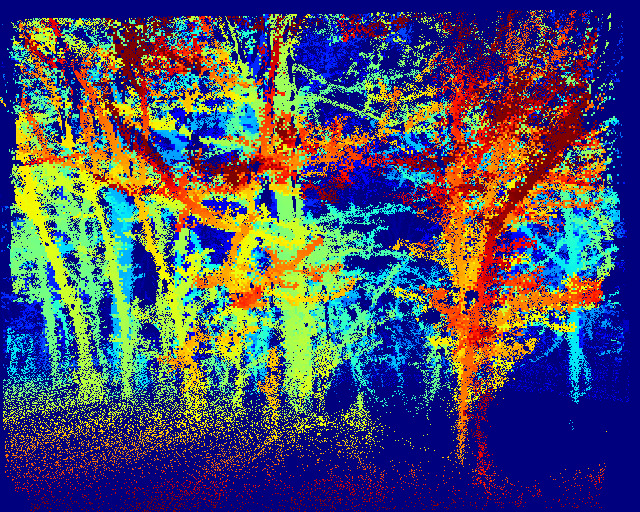} &
    \includegraphics[width=0.14\linewidth]{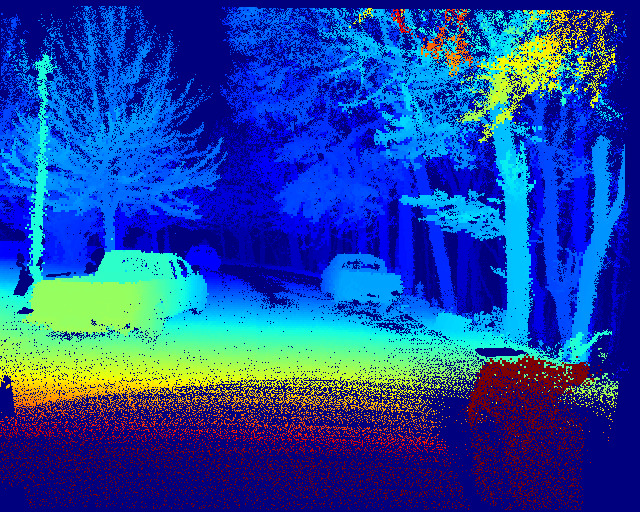} &
    \includegraphics[width=0.14\linewidth]{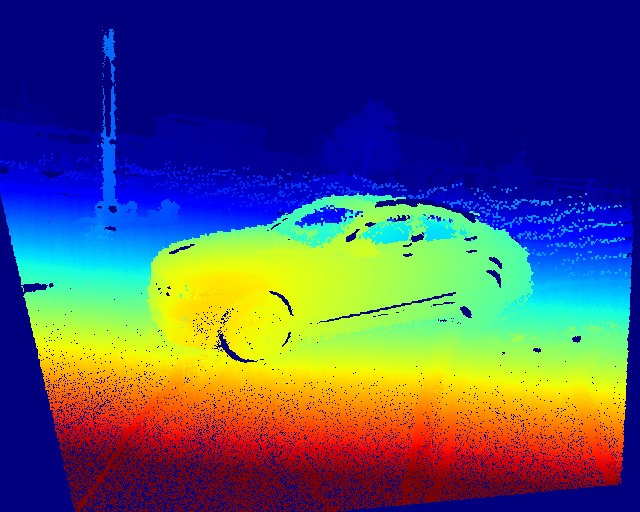} \\
    [4pt]
    Sparse Trees & Lamp Post & Single Tree & Park & Dense Trees & Night & Car \\
  \end{tabular}
  \caption{Data environment variety over all run locations. The variety of environments span urban settings, sparse trees, and dense trees.}
  \label{fig:data_variety}
\end{figure*}

%% file: text/04-evaluation.tex
\section{Evaluation}

\subsection{Thermal image pre-processing}
Thermal images captured in raw 16-bit format present challenges such as low texture, low contrast, and high noise patterns. To improve image quality and facilitate feature extraction for depth estimation, we apply several pre-processing steps to the stereo images:

1. \textbf{Intensity Binding} \cite{shin2023ms2}: Constrains the image histogram to the 1-99 percentile range to mitigate extreme outliers like the sun or smoke bombs, which can cause pixel saturation and detail loss.


2. \textbf{CLAHE (Contrast Limited Adaptive Histogram Equalization)} \cite{CLAHE}: Enhances local contrast and visibility without amplifying noise by adjusting tile histograms to a specified profile.


3. \textbf{Bilateral Filtering} \cite{BilateralFiltering}: Preserves edges and reduces noise by averaging pixel intensities with spatial and radiometric weights, maintaining edges in cluttered environments.

\input{figures/thermal_processing}

These preprocessing steps produce 8-bit images optimized for depth estimation models, characterized by enhanced contrast, preserved textures, and reduced noise as seen in \cref{fig:thermal_processing}. Lastly, we align the contrast between processed left and right stereo images $I_L, I_R$ by equalizing their mean intensity $\mu$ as shown in \cref{contrastAlign}.

\begin{equation}
\begin{cases}
    I_R = I_R \frac{\mu_{L}}{\mu_{R}} & \text{if } \mu_{R} < \mu_{L} \\
    I_L = I_L \frac{\mu_{R}}{\mu_{L}} & \text{if } \mu_{L} < \mu_{R}
\end{cases}
\label{contrastAlign}
\end{equation}

\subsection{Depth Model Selection}
We implemented 5 representative stereo depth estimation models to evaluate the capabilities of our new dataset in facilitating robust depth estimation for UAS navigation in cluttered environment.

\input{tables/results_supervised_2}

\textbf{Lightweight networks}: We selected 2 models optimized for rapid inference times, which is essential for deployment in sUAS where computational resources and response times are limited. \textbf{Fast-ACVNet}\cite{xu2023accuratefastacv} introduces Attention Concatenation Volume and Volume Attention Propagation for optimized cost volume construction and interpolation. \textbf{MobileStereoNet} \cite{shamsafar2022mobilestereonet} leveraged efficient point-wise and depth-wise convolutions to perform fast stereo matching on mobile platforms and its 2D version is used.   

\textbf{3D networks}: We selected 3 stereo depth model optimized for performance. \textbf{AANet} \cite{xu2020aanet} proposed sparse points-based intra-scale cost aggregation to address edge-fattening issues. \textbf{GWCNet} \cite{guo2019groupgwcnet} uses group-wise correlation to construct cost volume for improved feature similarities. \textbf{PSMNet} \cite{chang2018pyramidpsmnet} uses spatial pyramid for global context and 3D CNN to regularize cost volume. These models benchmark our dataset's depth prediction capabilities within a well-established framework.

\subsection{Loss and Evaluation Metrics}
Models are trained with a smooth L1 loss as implemented in their official source code. To evaluate the models, we utilize the commonly used evaluation metrics end-point error (EPE) and percentage of outliers. EPE calculates average absolute disparity error across all pixels, quantifying deviation of predicted disparities \(D_{\text{pred}}\) from their respective ground truth values \(D_{\text{GT}}\) as in Eq \ref{epe}. Here, \(N\) represents the total number of pixels.

\begin{equation}
\text{EPE} = \frac{1}{N} \sum_{i=1}^{N} |D_{\text{pred}, i} - D_{\text{GT}, i}|
\label{epe}
\end{equation}

The \(D1\) metric quantifies percentage of disparity outliers, effectively measuring the proportion of significant outliers in the predicted disparities \(D_{\text{pred}}\). Outliers are defined as pixels whose disparity errors are greater than $\max(3, 0.05 \cdot D_{\text{GT}})$. We also report the percentage of outliers greater than 1px , 2px and 3px.

%% file: figures/thermal_processing.tex
\begin{figure}[ht]
    \centering
    \renewcommand\arraystretch{0} 
    \setlength{\tabcolsep}{1pt} 

    \begin{subfigure}{0.15\textwidth}
        \includegraphics[width=\linewidth]{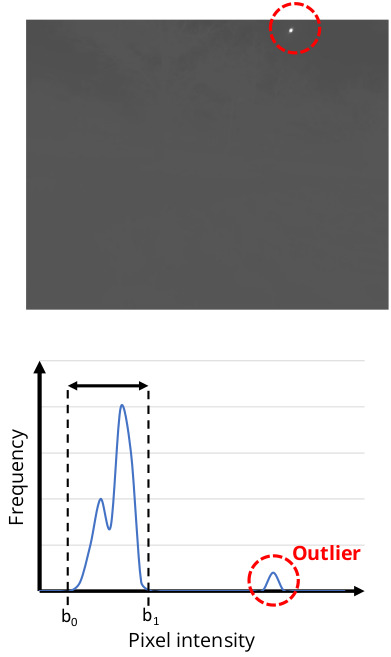}
        \caption{RAW image}
    \end{subfigure}
    \begin{subfigure}{0.15\textwidth}
        \includegraphics[width=\linewidth]{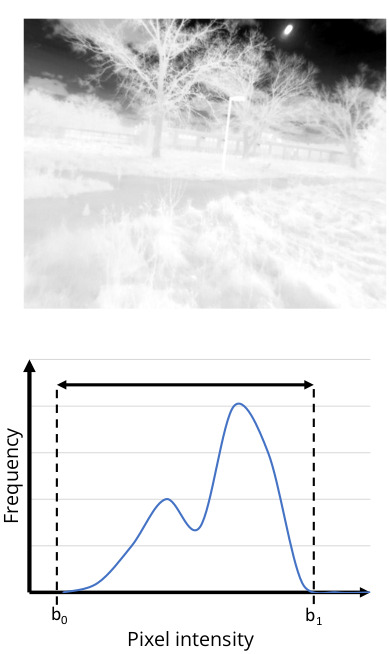}
        \caption{Intensity binding}
    \end{subfigure}
    \begin{subfigure}{0.15\textwidth}
        \includegraphics[width=\linewidth]{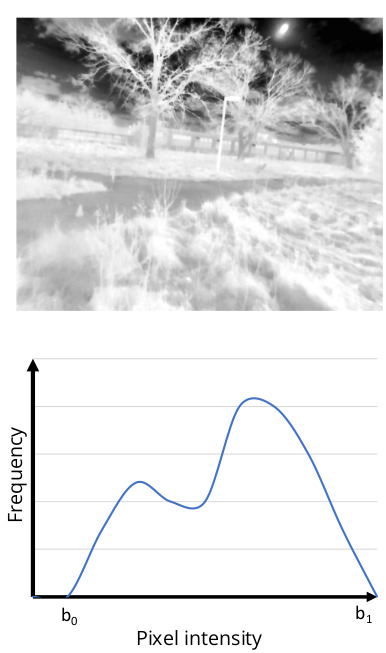}
        \caption{CLAHE + Bilateral}
    \end{subfigure}
    \caption{Effect of pre-processing on the raw 16-bit thermal image.}
    \label{fig:thermal_processing}
\end{figure}

%% file: tables/results_supervised_2.tex
\begin{table*}[t]
\centering
\vspace{8pt}
\caption{Quantitative results for disparity estimation on the test set of our FIReStereo-sUAS dataset. Lower is better.}
\label{tab:quant_results}
\resizebox{0.92\linewidth}{!}{%
\begin{tabular}{l|l|rrrrr|rr} 
\hline
\multirow{2}{*}{Methods}            & \multirow{2}{*}{Time/Condition}         & \multicolumn{5}{c|}{Error $\downarrow$}                                                                                                                                                                                                                                           & \multicolumn{2}{c}{Speed $\downarrow$}                                              \\ 
\cline{3-9}
                                    &                                         & \multicolumn{1}{c}{EPE(px)}                        & \multicolumn{1}{c}{D1(\%)}                        & \multicolumn{1}{c}{$>$1px(\%)}                        & \multicolumn{1}{c}{$>$2px(\%)}                        & \multicolumn{1}{c|}{$>$3px(\%)}                      & \multicolumn{1}{c}{FLOPS(G)}     & \multicolumn{1}{c}{Runtime(ms)}     \\ 
\hline
\multirow{3}{*}{PSMNet \cite{chang2018pyramidpsmnet}}             & Day/Cloudy                              & 1.260~~                                            & 8.10~~                                            & 27.44~~                                            & 13.17~~                                            & 8.10~~                                            & \multirow{3}{*}{462.6~~}         & \multirow{3}{*}{259.1~ ~~}          \\
                                    & Night/Rain                              & 1.671~~                                            & 12.42~~                                           & 32.80~~                                            & 17.76~~                                            & 12.42~~                                           &                                  &                                     \\
                                    & {\cellcolor[rgb]{0.753,0.753,0.753}}Avg & {\cellcolor[rgb]{0.753,0.753,0.753}}1.285~~        & {\cellcolor[rgb]{0.753,0.753,0.753}}8.35~~        & {\cellcolor[rgb]{0.753,0.753,0.753}}27.75~~        & {\cellcolor[rgb]{0.753,0.753,0.753}}13.44~~        & {\cellcolor[rgb]{0.753,0.753,0.753}}8.35~~        &                                  &                                     \\ 
\hline
\multirow{3}{*}{GwcNet-gc \cite{guo2019groupgwcnet}}          & Day/Cloudy                              & 1.250~~                                            & 7.96~~                                            & 26.46~~                                            & 12.66~~                                            & 7.96~~                                            & \multirow{3}{*}{436.6~~}         & \multirow{3}{*}{190.7~ ~~}          \\
                                    & Night/Rain                              & 1.610~~                                            & 11.63~~                                           & 30.01~~                                            & 16.38~~                                            & 11.63~~                                           &                                  &                                     \\
                                    & {\cellcolor[rgb]{0.753,0.753,0.753}}Avg & {\cellcolor[rgb]{0.753,0.753,0.753}}1.271~~        & {\cellcolor[rgb]{0.753,0.753,0.753}}\textbf{8.17}~~ & {\cellcolor[rgb]{0.753,0.753,0.753}}\textbf{26.66}~~ & {\cellcolor[rgb]{0.753,0.753,0.753}}\textbf{12.88}~~ & {\cellcolor[rgb]{0.753,0.753,0.753}}\textbf{8.18}~~ &                                  &                                     \\ 
\hline
\multirow{3}{*}{AANet \cite{xu2020aanet}}              & Day/Cloudy                              & 1.245~~                                            & 8.03~~                                            & 26.82~~                                            & 12.80~~                                            & 8.04~~                                            & \multirow{3}{*}{143.6~~}         & \multirow{3}{*}{152.5~ ~~}          \\
                                    & Night/Rain                              & 1.585~~                                            & 11.55~~                                           & 30.61~~                                            & 16.48~~                                            & 11.54~~                                           &                                  &                                     \\
                                    & {\cellcolor[rgb]{0.753,0.753,0.753}}Avg & {\cellcolor[rgb]{0.753,0.753,0.753}}\textbf{1.266}~~ & {\cellcolor[rgb]{0.753,0.753,0.753}}8.24~~        & {\cellcolor[rgb]{0.753,0.753,0.753}}27.06~~        & {\cellcolor[rgb]{0.753,0.753,0.753}}13.03~~        & {\cellcolor[rgb]{0.753,0.753,0.753}}8.25~~        &                                  &                                     \\ 
\hline
\multirow{3}{*}{MobileStereoNet-2D \cite{shamsafar2022mobilestereonet}} & Day/Cloudy                              & 1.888~~                                            & 13.70~~                                           & 42.52~~                                            & 21.65~~                                            & 13.71~~                                           & \multirow{3}{*}{77.9~~}          & \multirow{3}{*}{120.7~ ~~}          \\
                                    & Night/Rain                              & 2.441~~                                            & 20.17~~                                           & 52.64~~                                            & 29.82~~                                            & 20.17~~                                           &                                  &                                     \\
                                    & {\cellcolor[rgb]{0.753,0.753,0.753}}Avg & {\cellcolor[rgb]{0.753,0.753,0.753}}1.922~~        & {\cellcolor[rgb]{0.753,0.753,0.753}}14.10~~       & {\cellcolor[rgb]{0.753,0.753,0.753}}43.11~~        & {\cellcolor[rgb]{0.753,0.753,0.753}}22.15~~        & {\cellcolor[rgb]{0.753,0.753,0.753}}14.11~~       &                                  &                                     \\ 
\hline
\multirow{3}{*}{Fast-ACVNet \cite{xu2023accuratefastacv}}        & Day/Cloudy                              & 1.244~~                                            & 8.08~~                                            & 26.76~~                                            & 12.76~~                                            & 8.08~~                                            & \multirow{3}{*}{\textbf{39.5~~}} & \multirow{3}{*}{\textbf{42.0~~~~}}  \\
                                    & Night/Rain                              & 1.599~~                                            & 11.75~~                                           & 30.74~~                                            & 16.70~~                                            & 11.75~~                                           &                                  &                                     \\
                                    & {\cellcolor[rgb]{0.753,0.753,0.753}}Avg & {\cellcolor[rgb]{0.753,0.753,0.753}}\textbf{1.266}~~ & {\cellcolor[rgb]{0.753,0.753,0.753}}8.30~~        & {\cellcolor[rgb]{0.753,0.753,0.753}}27.01~~        & {\cellcolor[rgb]{0.753,0.753,0.753}}13.00~~        & {\cellcolor[rgb]{0.753,0.753,0.753}}8.30~~        &                                  &                                     \\
\hline
\end{tabular}
}
\end{table*}

%% file: text/05-results.tex
\section{RESULTS}

\subsection{Implementation details}
All models are trained until convergence using the architecture provided in the official source code repository. We employ the Adam optimizer for gradient descent with a learning rate of $1e^{-4}$. Training is conducted on 4x NVIDIA H100 80GB GPUs. The input image size is fixed at 640x512 for all models, with a batch size of 16. Networks are initialized with the provided backbone feature extractor model, following their original implementations \cite{xu2023accuratefastacv, xu2020aanet, guo2019groupgwcnet, shamsafar2022mobilestereonet}.

\subsection{Quantitative results}
We report the quantitative metrics on the test set for the selected representative models
analyzed separately for the day/cloudy and night/rainy conditions. For computational performance, we measure the Floating Point Operations Per Second (FLOPS) in billions and average inference time in milliseconds at a 640x512 resolution on an NVIDIA RTX 3060 GPU. 
Results are shown in Table \ref{tab:quant_results}.

We observe that all models perform better in day/cloudy conditions compared to night/rainy conditions. The best average EPE score was 1.27px from AANet and Fast-ACVNet. GwcNet-gc performs best in D1 and the 1,2,3 px outlier errors, with a D1 score of 8.17\%. However, GwcNet-gc has 6.91M parameters, while PSMNet, AANet, Fast-ACVNet, MobileStereoNet-2D have 5.23M, 4M, 3.2M, and 2.35M parameters respectively. 

We observe that the lightweight model, Fast-ACVNet, with an inference speed of 42ms and a FLOPS of 39.5 G, delivers performance comparable to the representative GwcNet-gc and PSMNet model. Therfore, we select Fast-ACVNet for the qualitative results, as it is best suited for running on a low Size, Weight, Power, and Cost (SWaP-C) system while maintaining similar performance to the more resource-intensive models. 

\subsection{Qualitative results}
We report the qualitative inference results on the test set and compare the disparity prediction with ground truth. Additionally, we also run model inference on unseen smoke-filled images to show model generalization.

\cref{fig:qualitative_test} shows the qualitative results on the test set. We evaluate the model in day and night conditions. We observe that the model can estimate the disparity for challenging objects such as thin tree branches and poles. We also note that the disparity is accurately estimated for nearby as well as far obstacles.

We further evaluate the trained model on unseen environment with highly dense smoke conditions. The results are shown in \cref{fig:qualitative_smoke}. In \cref{fig:qualitative_smoke}(a) the smoke bomb is seen as a distinct hotspot and there is a trail of hot smoke near it. We also see a faint heat signature of the smoke bomb behind the turned car in \cref{fig:qualitative_smoke}(e). We observe that the thermal cameras are able to see through most of the smoke. The model is able to correctly estimate the obstacle disparity for these conditions. We can thus infer that the models trained with our training set generalize to smoke-filled environments.

\input{figures/qualitative_test}

%% file: figures/qualitative_test.tex
\begin{figure*}[ht]
\vspace{12pt}
\centering

\parbox[t]{0.06\textwidth}{\small Day/ \\Close}%
\begin{minipage}{0.95\textwidth}
\begin{subfigure}{0.16\textwidth}
    \includegraphics[width=\linewidth]{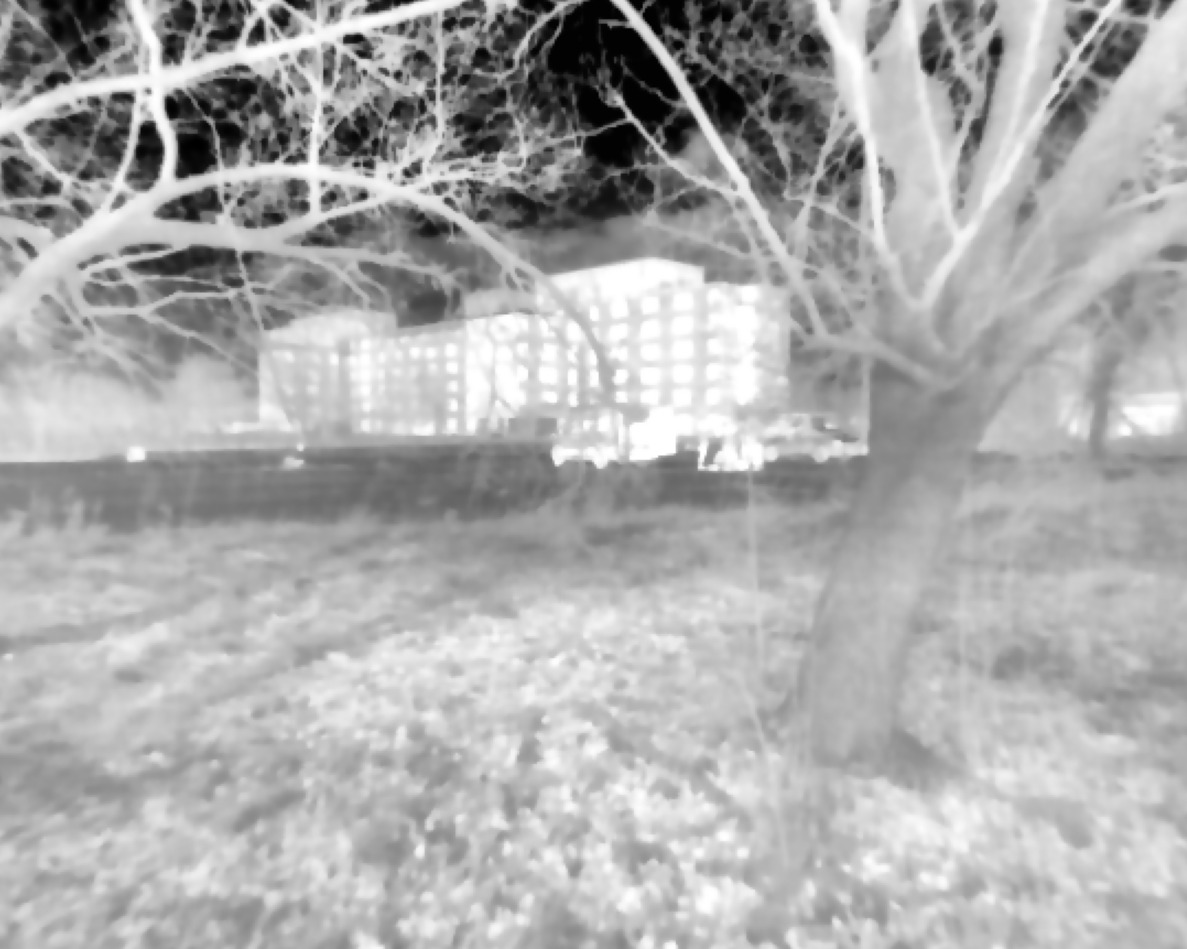}
\end{subfigure}
\begin{subfigure}{0.16\textwidth}
    \includegraphics[width=\linewidth]{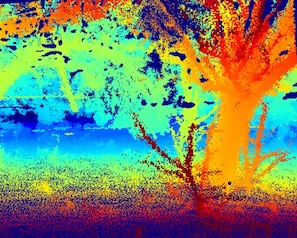}
\end{subfigure}
\begin{subfigure}{0.16\textwidth}
    \includegraphics[width=\linewidth]{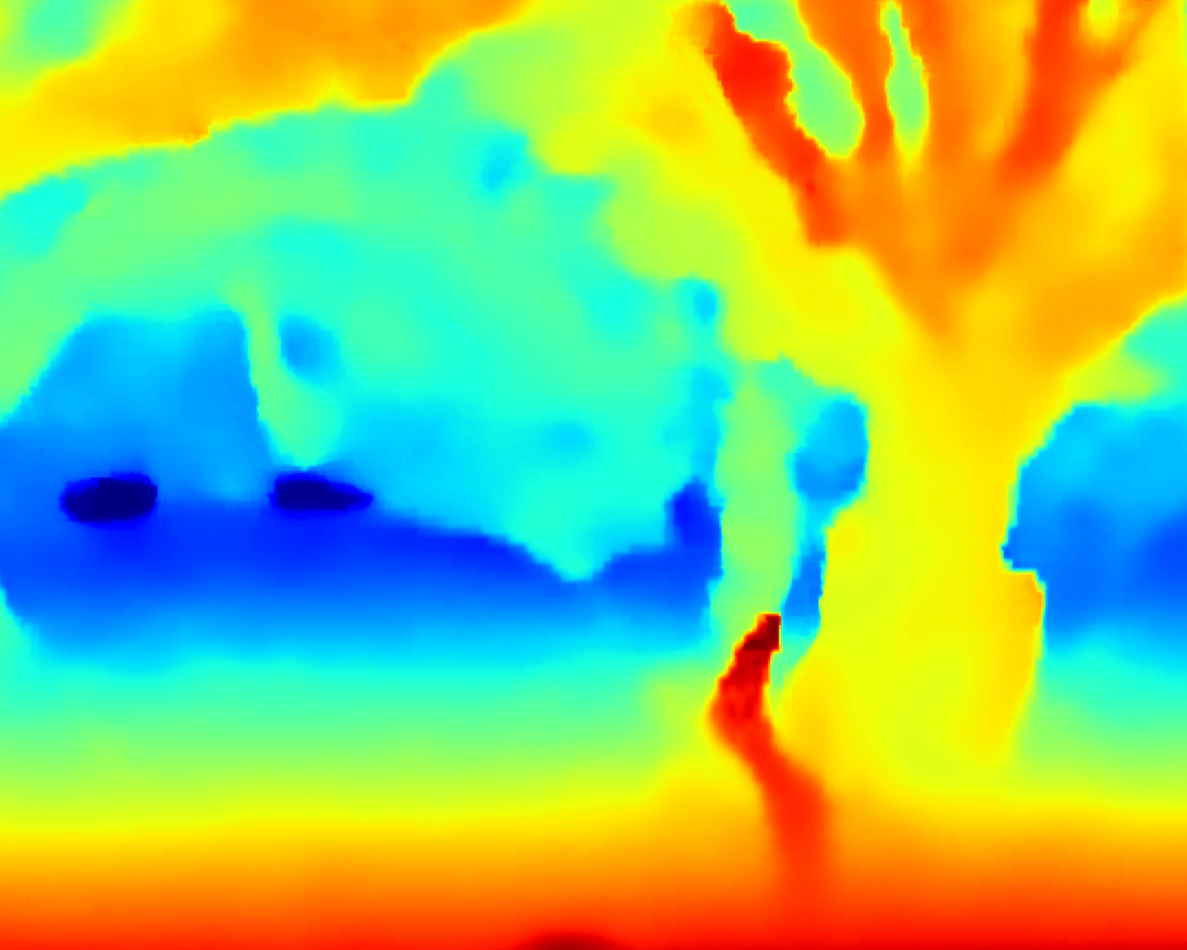}
\end{subfigure}
\hspace{0.6mm}
\begin{subfigure}{0.16\textwidth}
    \includegraphics[width=\linewidth]{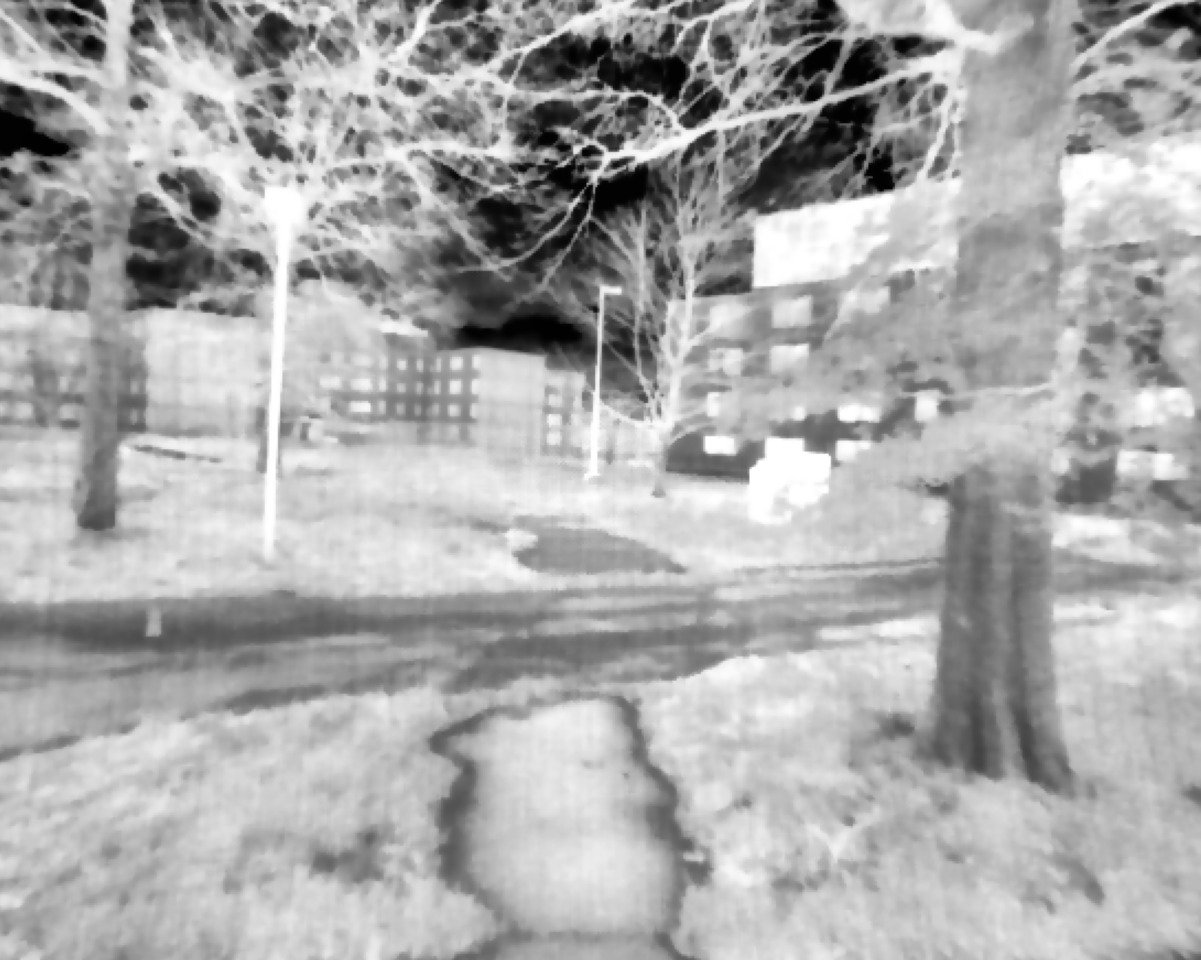}
\end{subfigure}
    \begin{subfigure}{0.16\textwidth}
    \includegraphics[width=\linewidth]{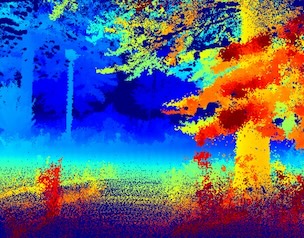}
\end{subfigure}
    \begin{subfigure}{0.16\textwidth}
    \includegraphics[width=\linewidth]{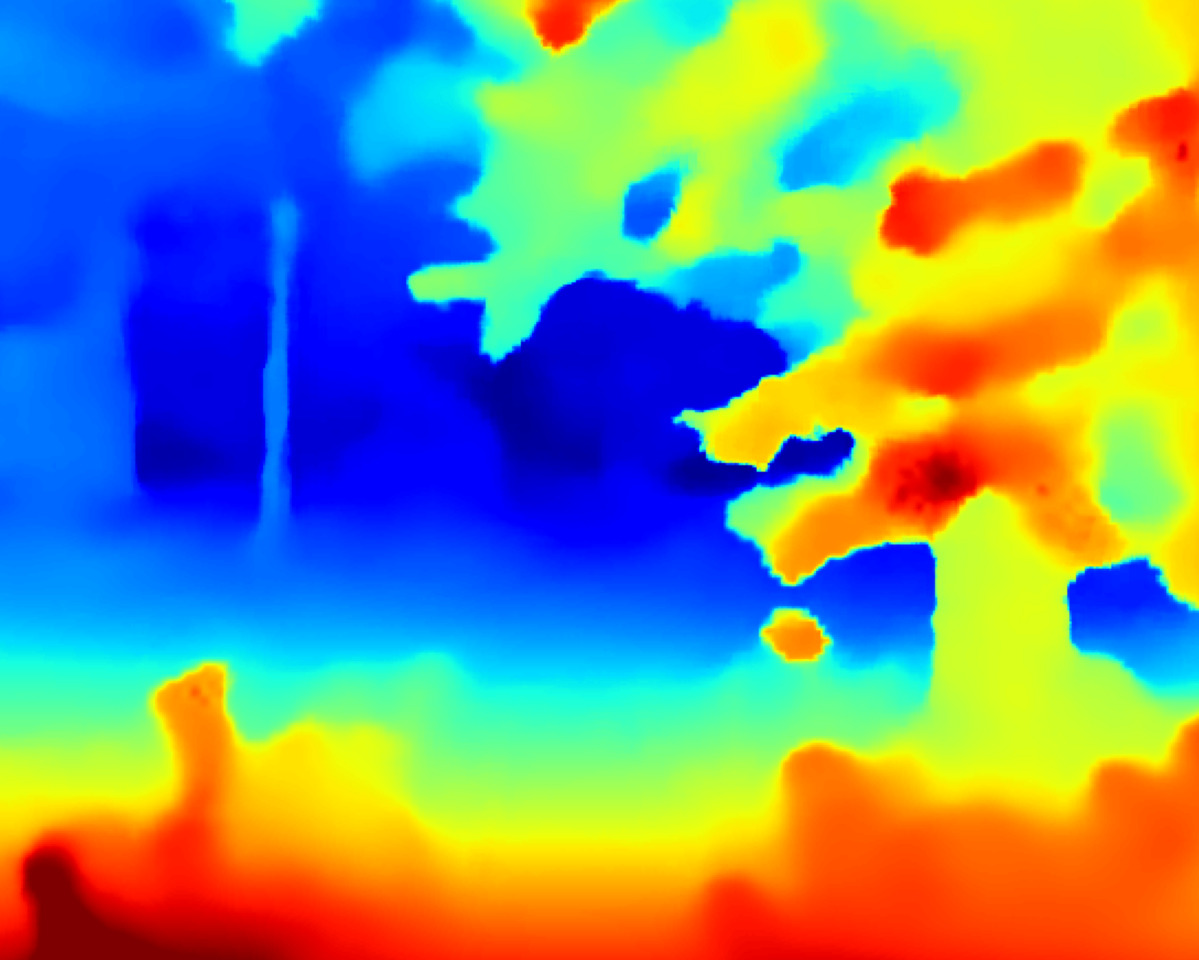}
\end{subfigure}
\end{minipage}
\vspace{0.6mm}

\parbox[t]{0.06\textwidth}{\small Day/ \\Far}%
\begin{minipage}{0.95\textwidth}
\begin{subfigure}{0.16\textwidth}
    \includegraphics[width=\linewidth]{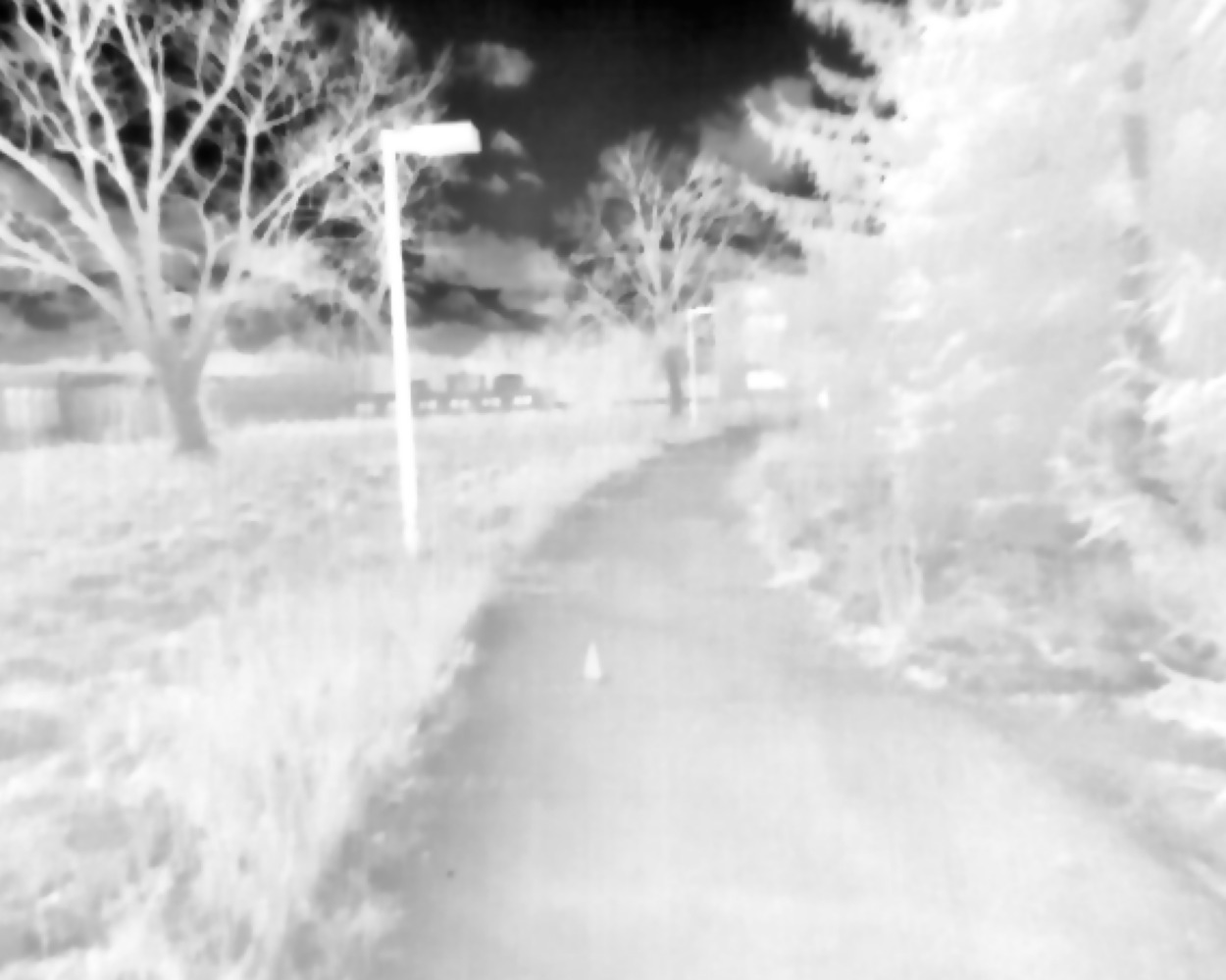}
\end{subfigure}
\begin{subfigure}{0.16\textwidth}
    \includegraphics[width=\linewidth]{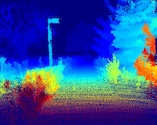}
\end{subfigure}
\begin{subfigure}{0.16\textwidth}
    \includegraphics[width=\linewidth]{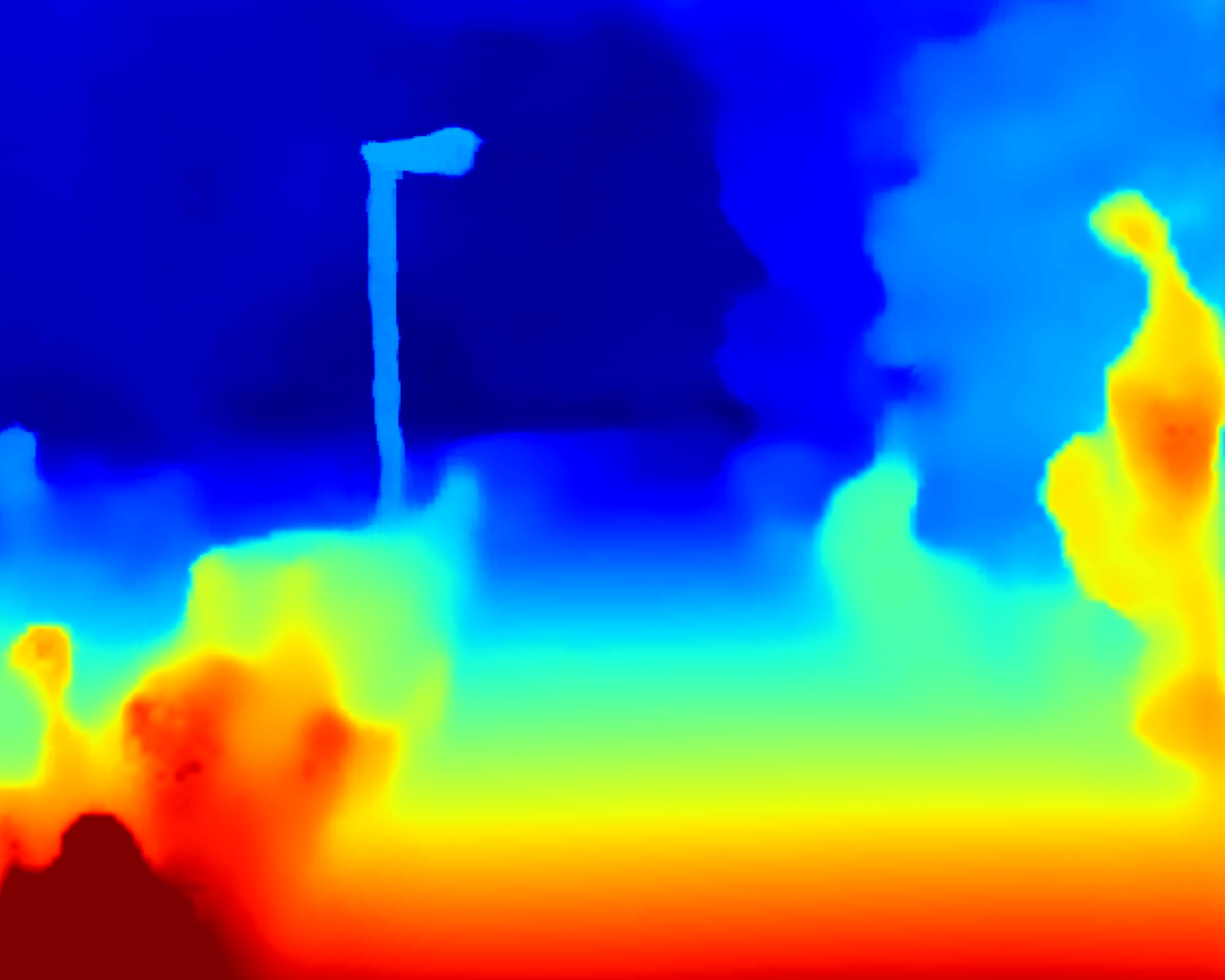}
\end{subfigure}
\hspace{0.8mm}
\begin{subfigure}{0.16\textwidth}
    \includegraphics[width=\linewidth]{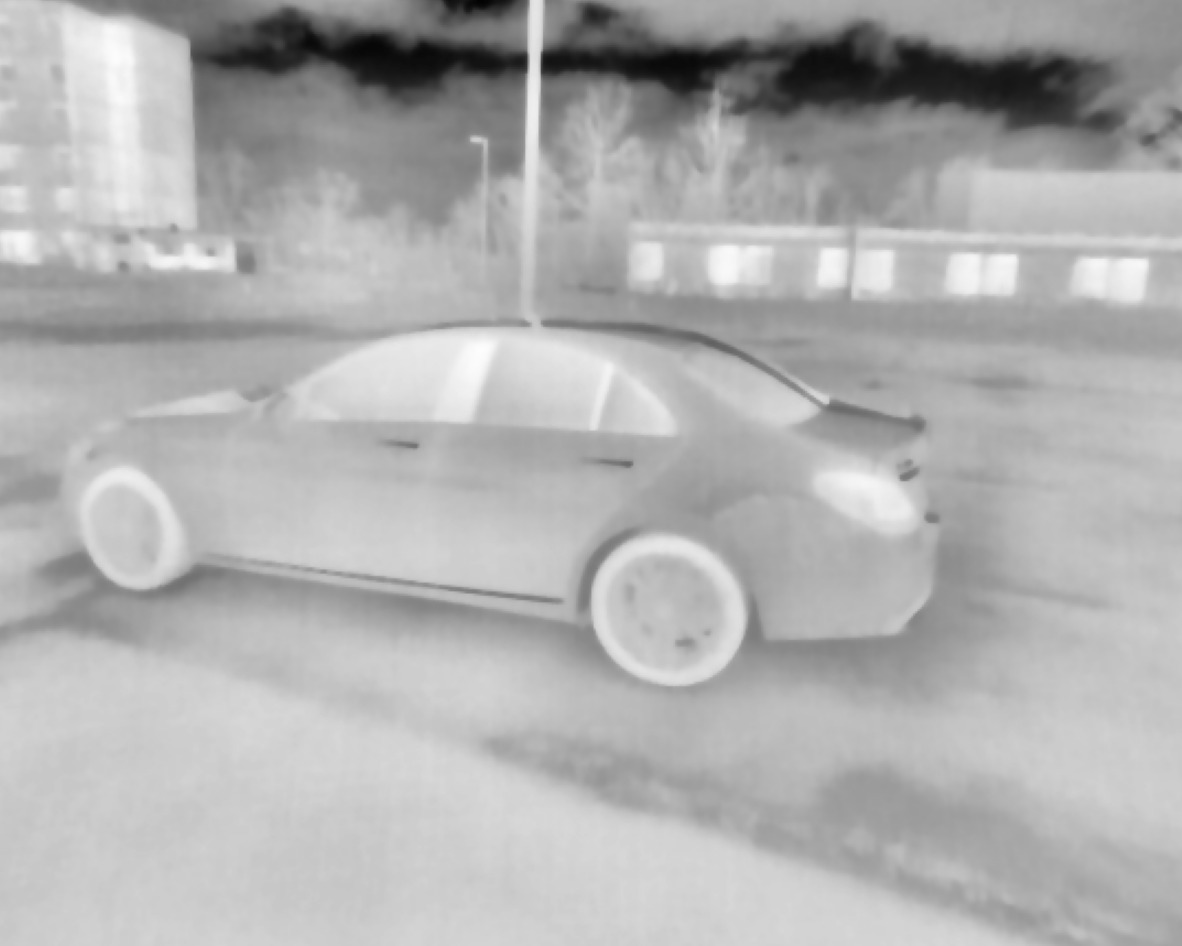}
\end{subfigure}
\begin{subfigure}{0.16\textwidth}
    \includegraphics[width=\linewidth]{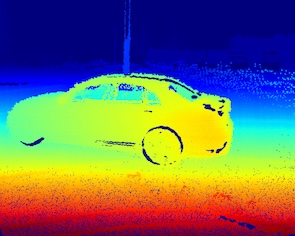}
\end{subfigure}
\begin{subfigure}{0.16\textwidth}
    \includegraphics[width=\linewidth]{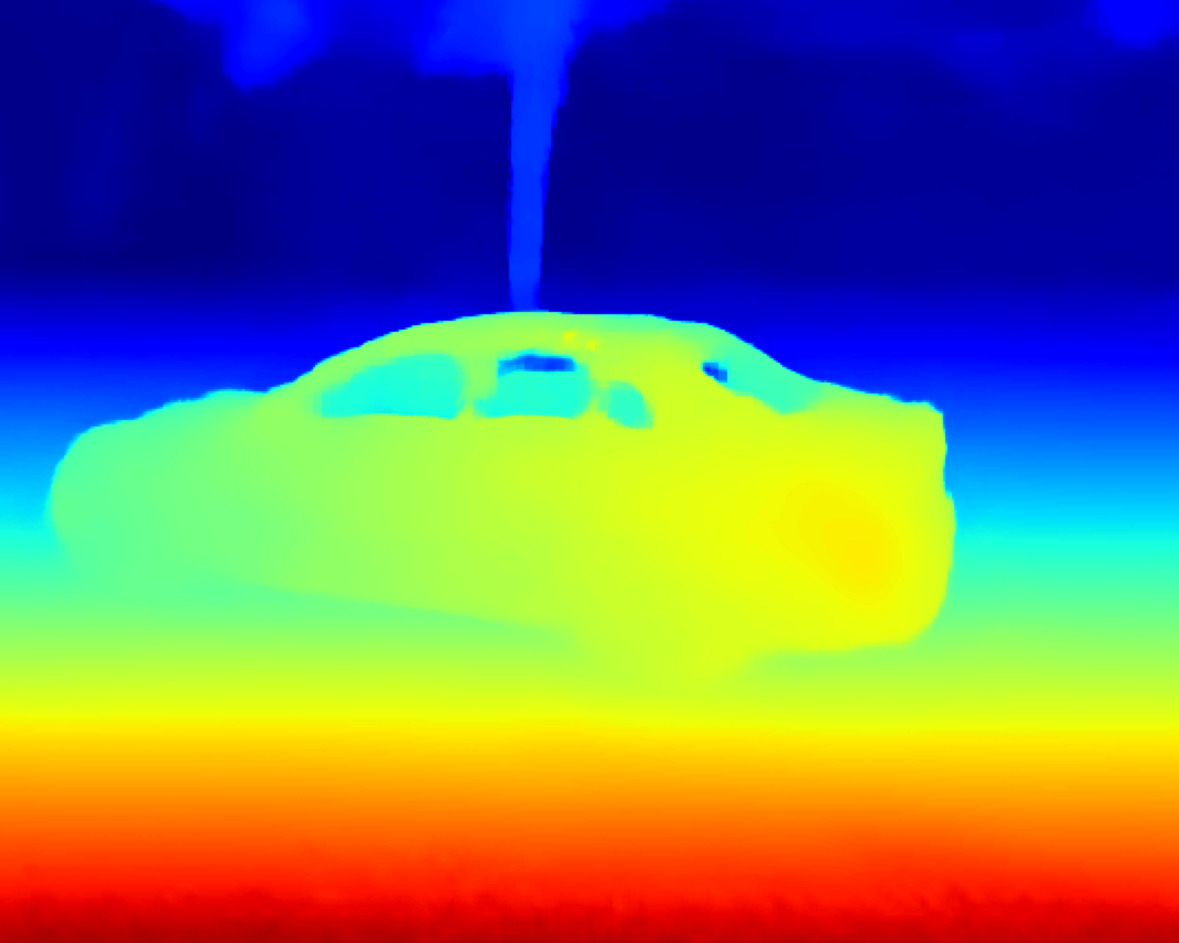}
\end{subfigure}
\end{minipage}
\vspace{0.6mm}

\parbox[t]{0.06\textwidth}{\small Night/ \\Close}%
\begin{minipage}{0.95\textwidth}
\begin{subfigure}{0.16\textwidth}
    \includegraphics[width=\linewidth]{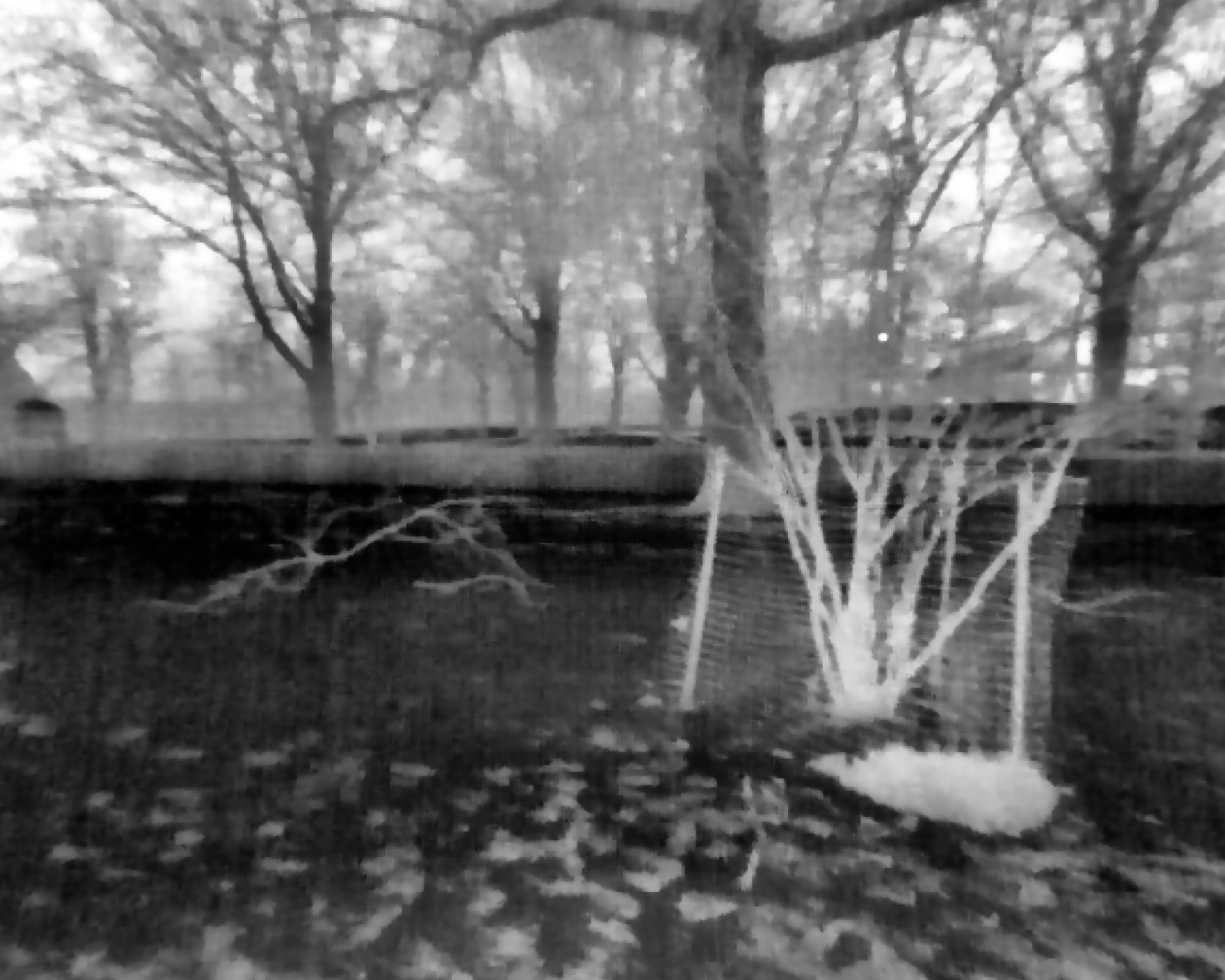}
\end{subfigure}
\begin{subfigure}{0.16\textwidth}
    \includegraphics[width=\linewidth]{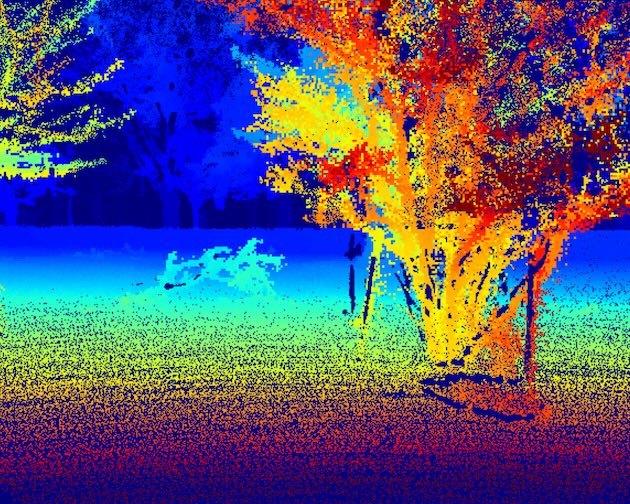}
\end{subfigure}
\begin{subfigure}{0.16\textwidth}
    \includegraphics[width=\linewidth]{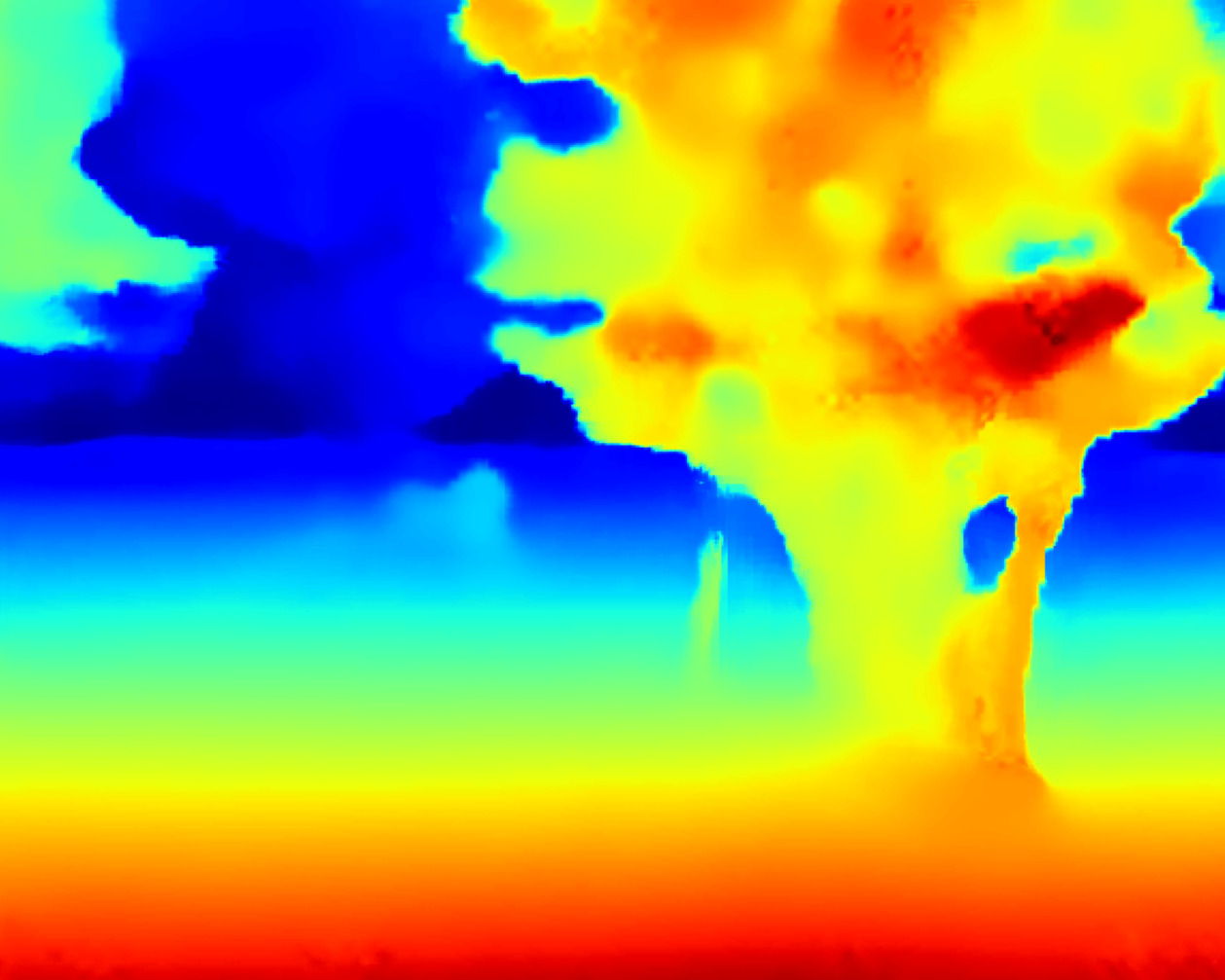}
\end{subfigure}
\hspace{0.8mm}
\begin{subfigure}{0.16\textwidth}
    \includegraphics[width=\linewidth]{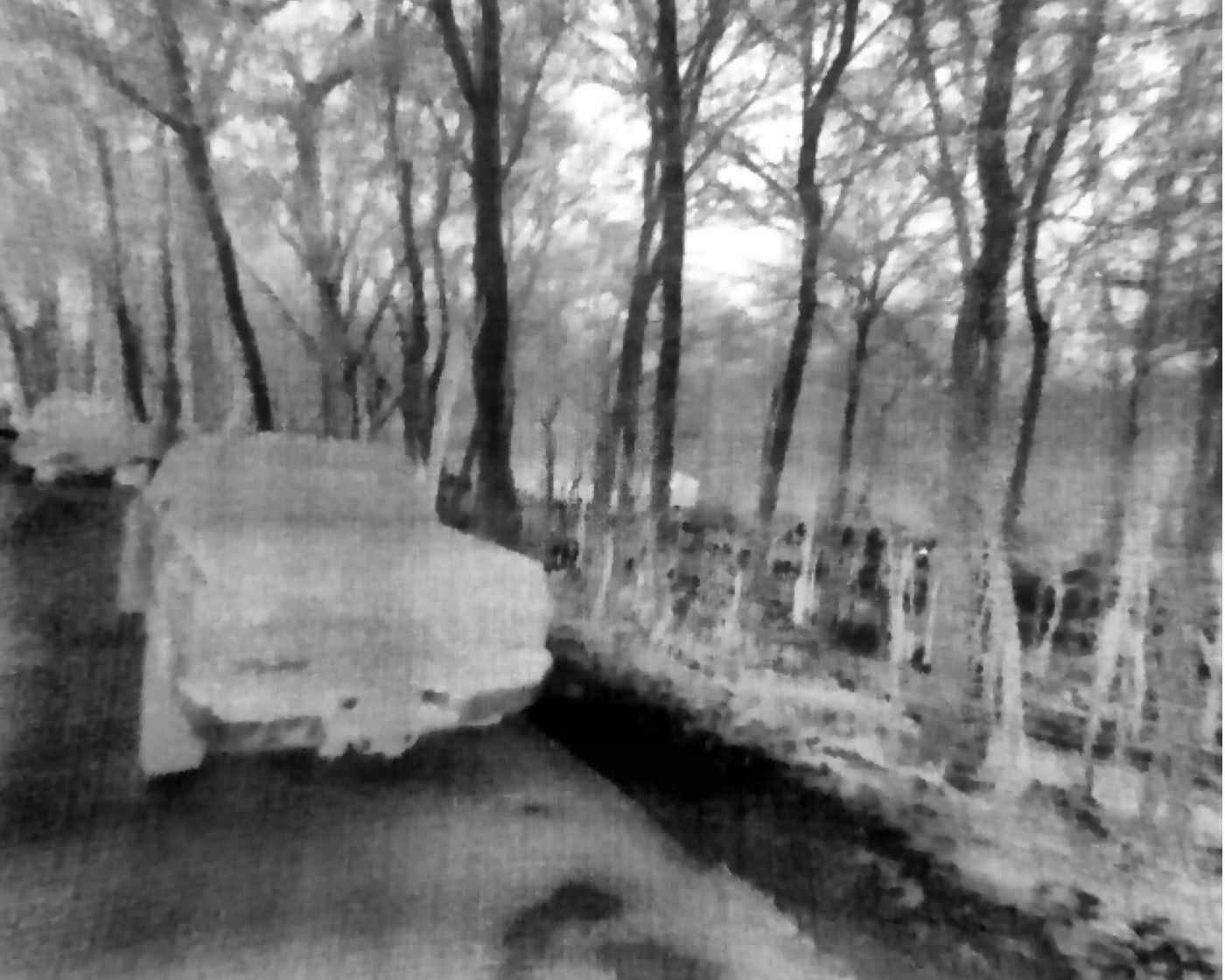}
\end{subfigure}
\begin{subfigure}{0.16\textwidth}
    \includegraphics[width=\linewidth]{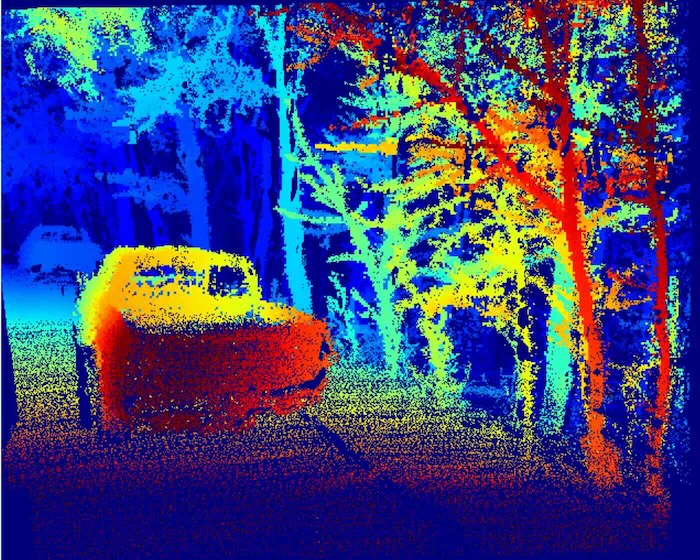}
\end{subfigure}
\begin{subfigure}{0.16\textwidth}
    \includegraphics[width=\linewidth]{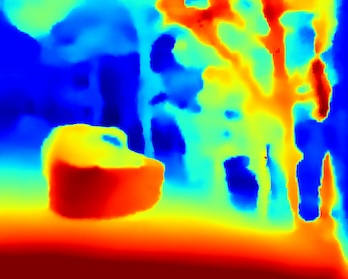}
\end{subfigure}
\end{minipage}
\vspace{0.6mm}

\parbox[t]{0.06\textwidth}{\small Night/ \\Far}%
\begin{minipage}{0.95\textwidth}
\begin{subfigure}{0.16\textwidth}
    \includegraphics[width=\linewidth]{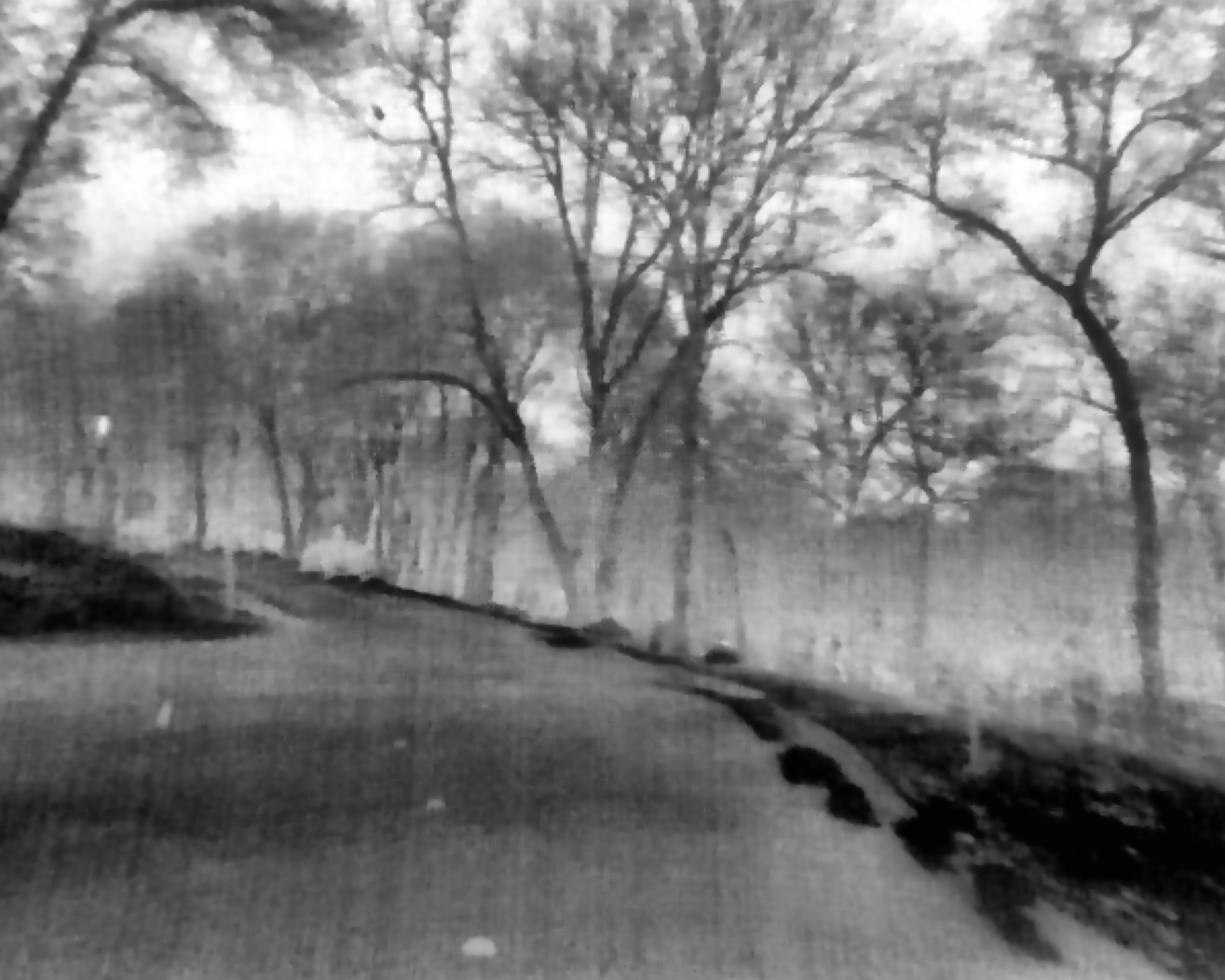}
    \caption{Thermal}
\end{subfigure}
\begin{subfigure}{0.16\textwidth}
    \includegraphics[width=\linewidth]{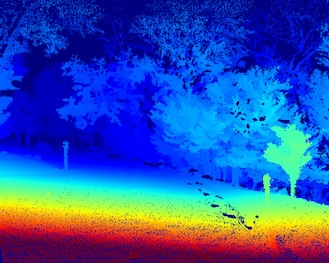}
    \caption{Ground Truth}
\end{subfigure}
\begin{subfigure}{0.16\textwidth}
    \includegraphics[width=\linewidth]{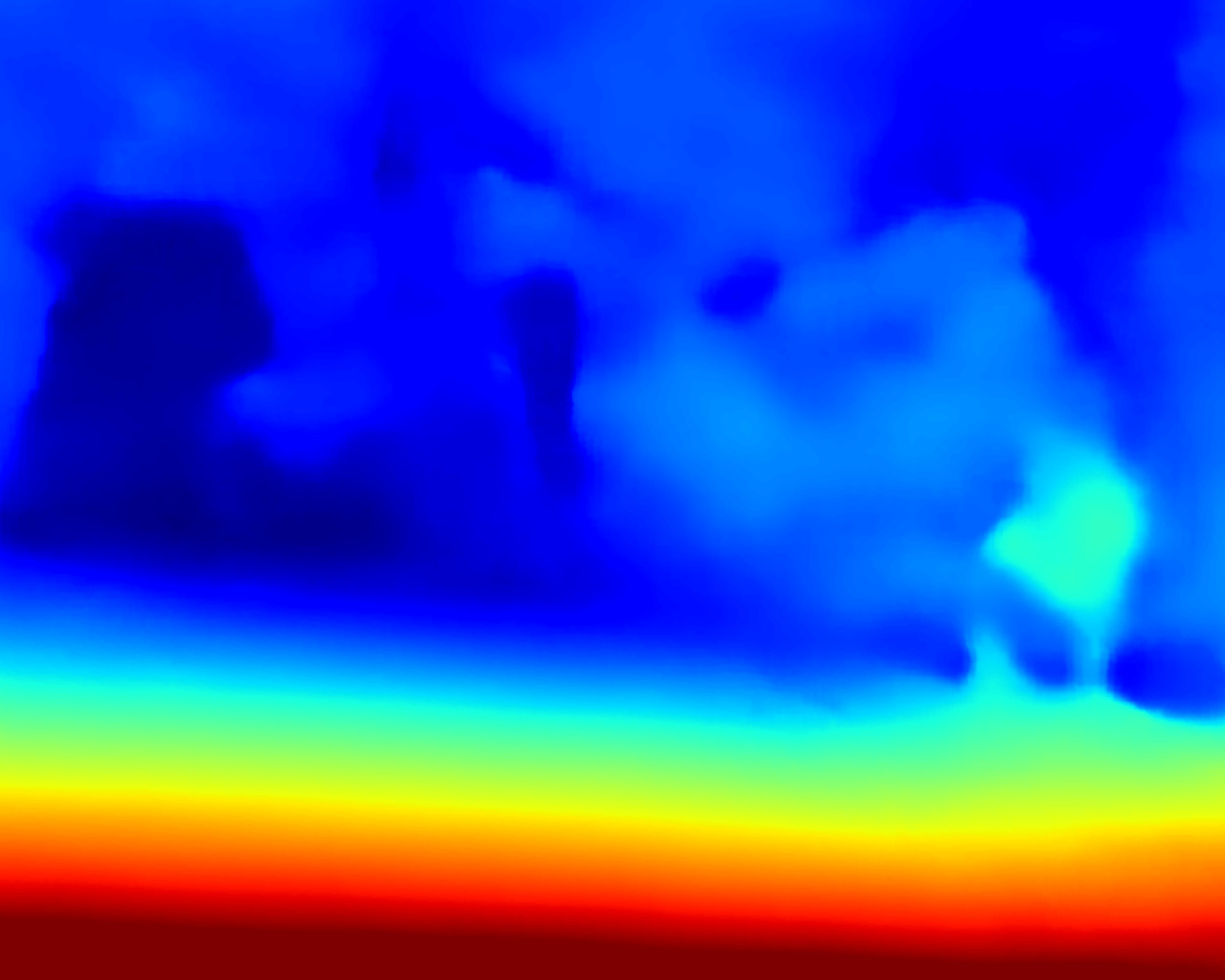}
    \caption{Estimated}
\end{subfigure}
\hspace{0.8mm}
\begin{subfigure}{0.16\textwidth}
    \includegraphics[width=\linewidth]{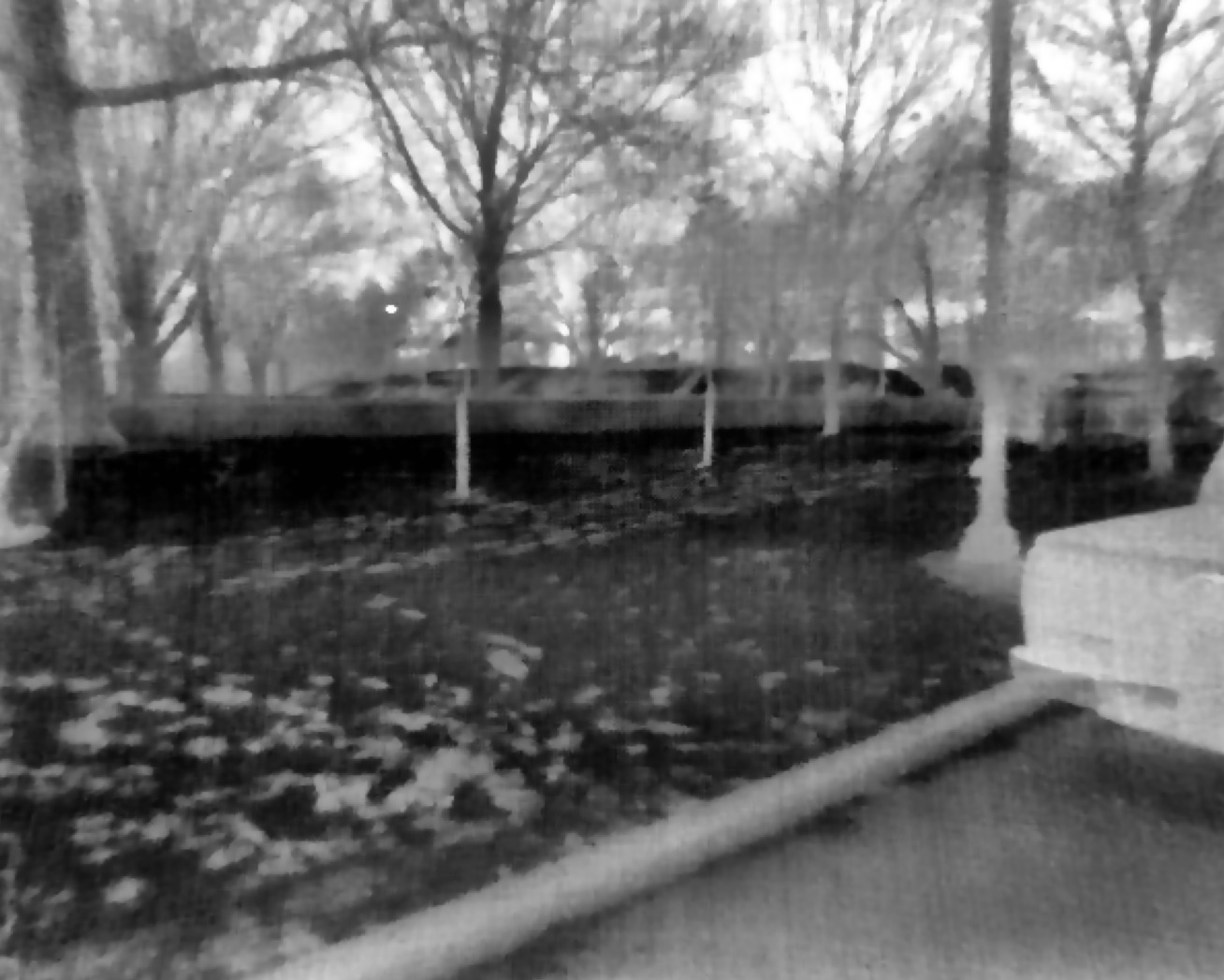}
    \caption{Thermal}
\end{subfigure}
\begin{subfigure}{0.16\textwidth}
    \includegraphics[width=\linewidth]{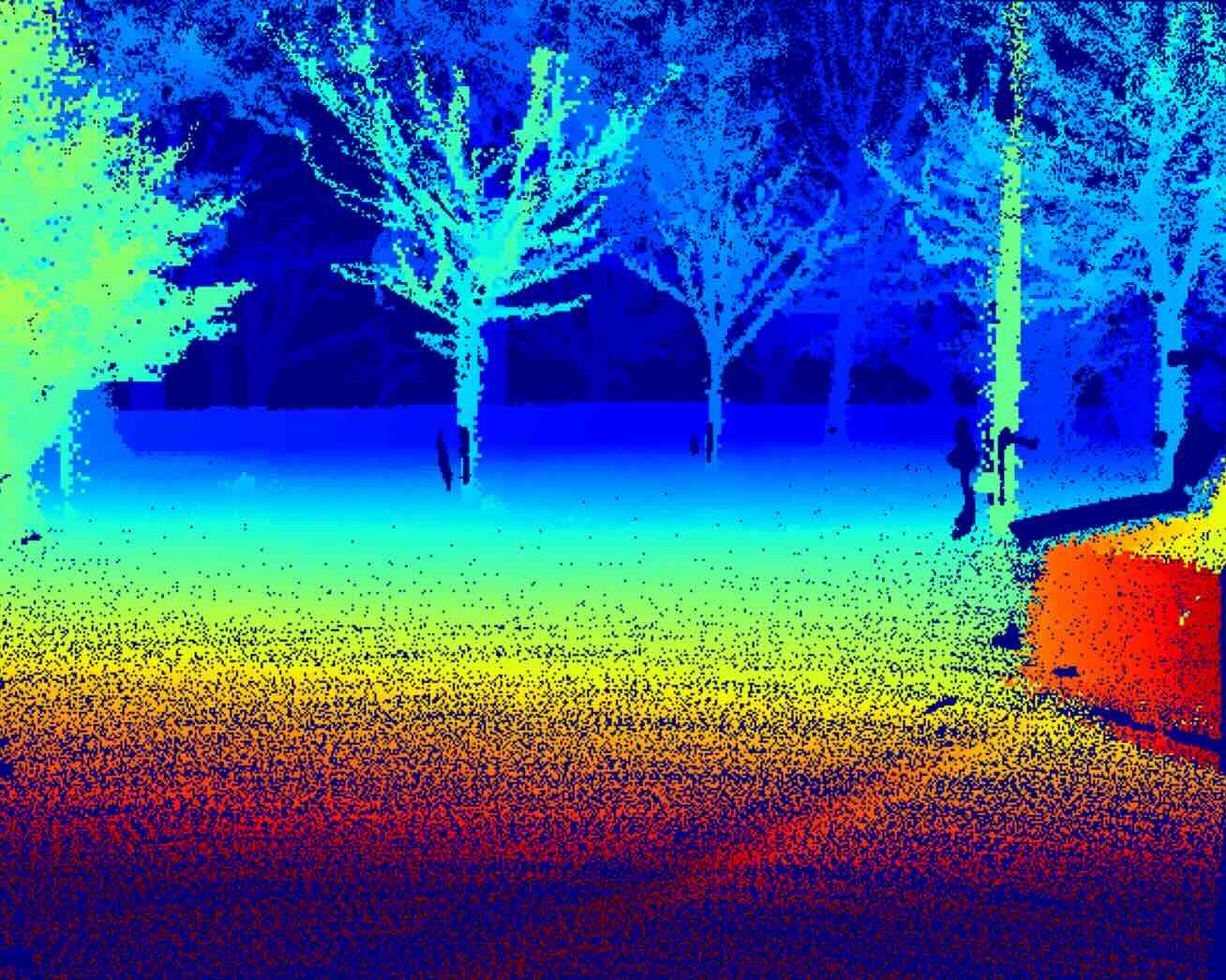}
    \caption{Ground Truth}
\end{subfigure}
\begin{subfigure}{0.16\textwidth}
    \includegraphics[width=\linewidth]{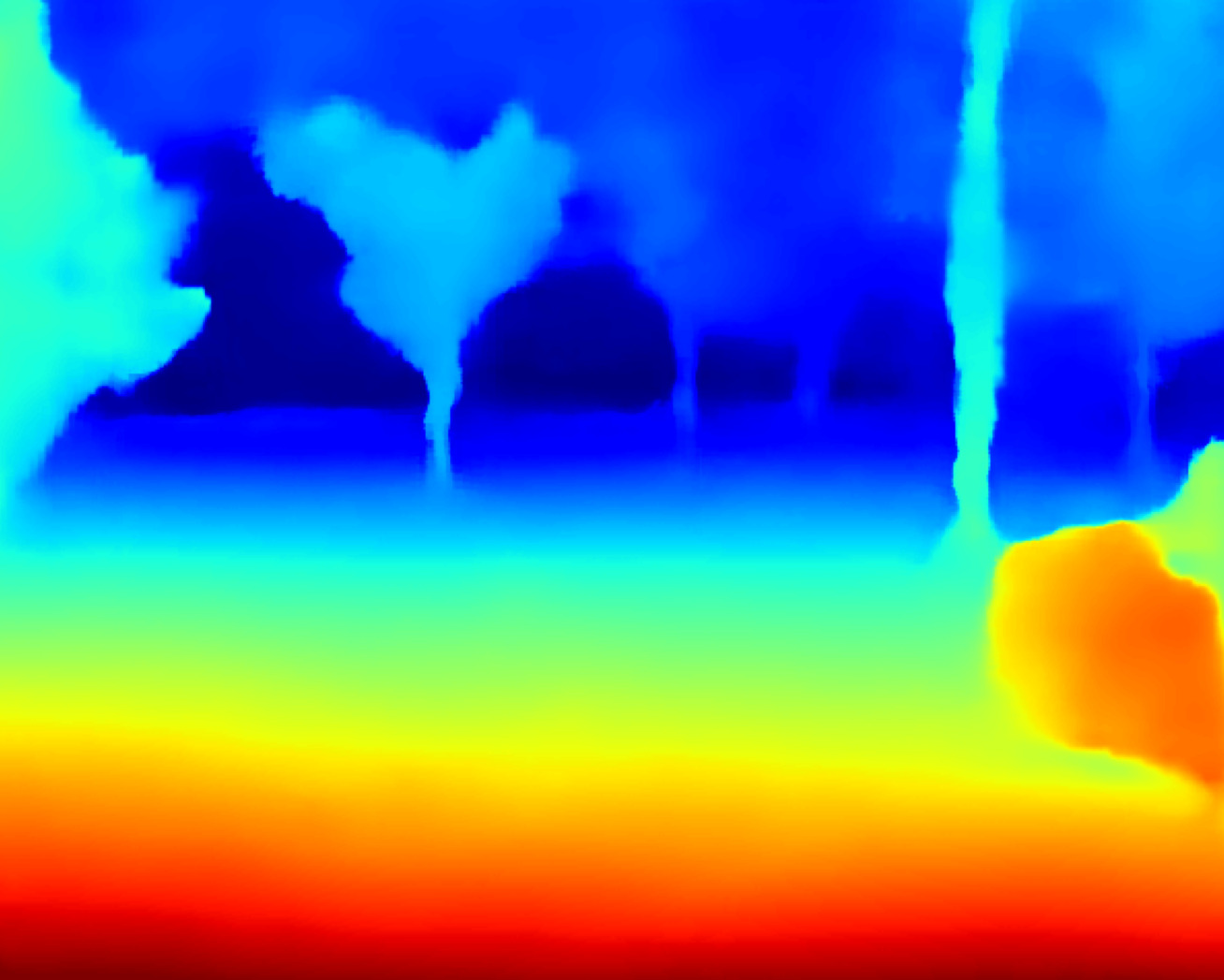}
    \caption{Estimated}
\end{subfigure}
\end{minipage}
\caption{Disparity estimation result with Fast-ACVNet on test set.}
\vspace{-4pt}
\label{fig:qualitative_test}
\end{figure*}

%% file: text/06-discussion.tex
\section{DISCUSSION}

A critical aspect of ensuring accurate depth projection in our system is camera and LiDAR calibration. While we employed some manual corrections to address calibration challenges, future work could benefit from automated LiDAR-Thermal cross-calibration tools designed to handle the specific properties of thermal cameras.

For future research, we suggest designing more lightweight stereo matching networks that reduce the computational burden in sUAS, such as exploring efficient network architectures, model distillation, or quantization to create models that can run effectively on embedded systems. Another important consideration for safe navigation in the cluttered environments presented in the dataset is accurate detection of thin obstacles like branches. While bilateral filtering has been used in our study to preserve edges, incorporating semantic segmentation could further preserve fine structures.

The narrow stereo baseline characteristic of \emph{FIReStereo} is advantageous for developing perception algorithms tailored for compact outdoor robots. Meanwhile, it presents unique challenges in depth estimation, particularly for small or \input{figures/qualitative_unseen}
distant objects, which might require accuracy at sub-pixel level.

Future work could focus on enhancing sub-pixel stereo matching techniques, such as refining cost aggregation methods, performing image super-resolution, or incorporating post-processing steps like disparity refinement. 
Additionally, monocular-stereo cue fusion is a promising direction, where monocular depth cues are used when stereo reliability is limited due to the small baseline or objects are beyond the effective stereo range.

%% file: figures/qualitative_unseen.tex
\begin{figure}[ht!]
    \centering
    \begin{minipage}{0.9\columnwidth}
    \centering
    \begin{subfigure}{0.31\columnwidth}
        \includegraphics[width=\linewidth]{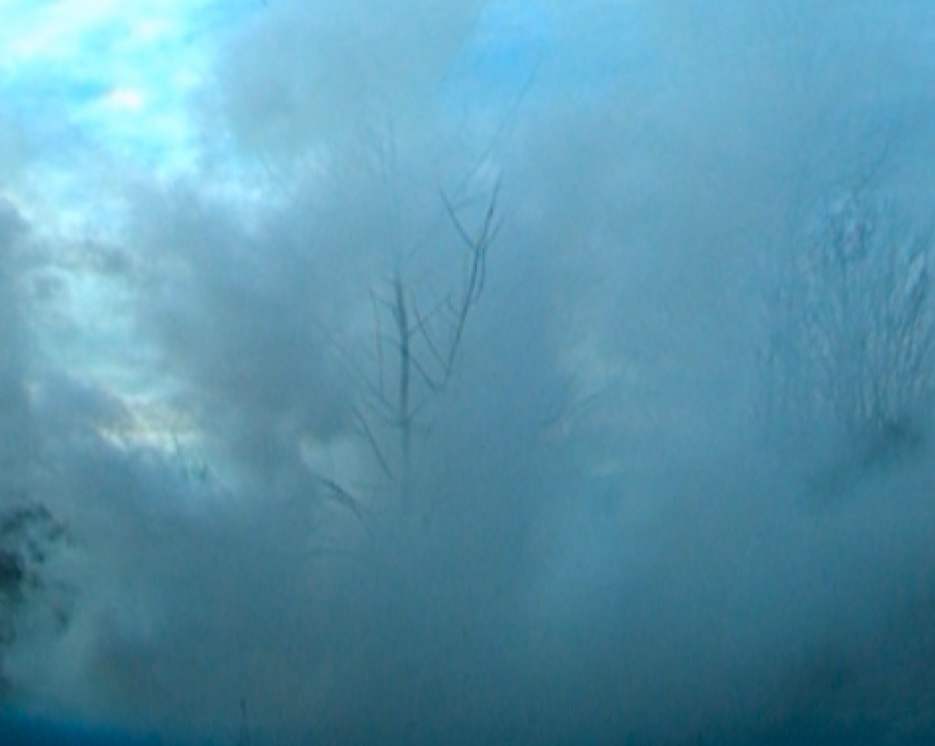}
    \end{subfigure}%
    \hspace{0.02\columnwidth}%
    \begin{subfigure}{0.31\columnwidth}
        \includegraphics[width=\linewidth]{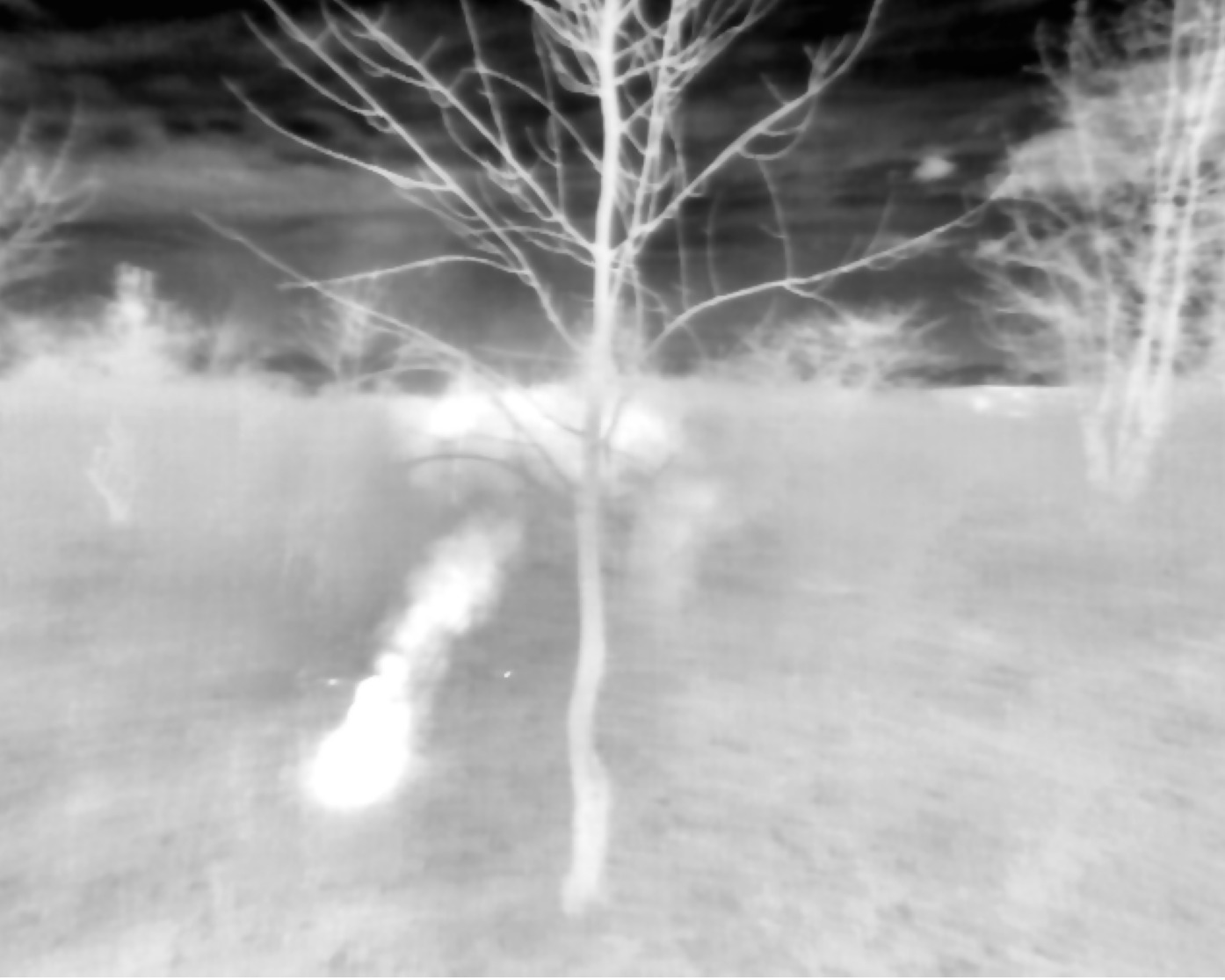}
    \end{subfigure}%
    \hspace{0.02\columnwidth}%
    \begin{subfigure}{0.31\columnwidth}
        \includegraphics[width=\linewidth]{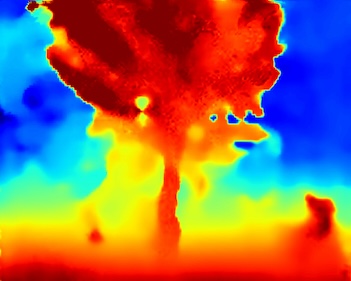}
    \end{subfigure}

    \vspace{0.4mm}

    \begin{subfigure}{0.31\columnwidth}
        \includegraphics[width=\linewidth]{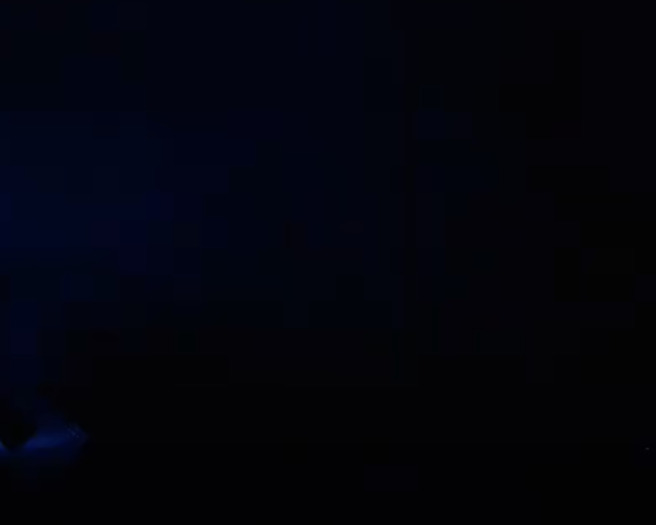}
    \end{subfigure}%
    \hspace{0.02\columnwidth}%
    \begin{subfigure}{0.31\columnwidth}
        \includegraphics[width=\linewidth]{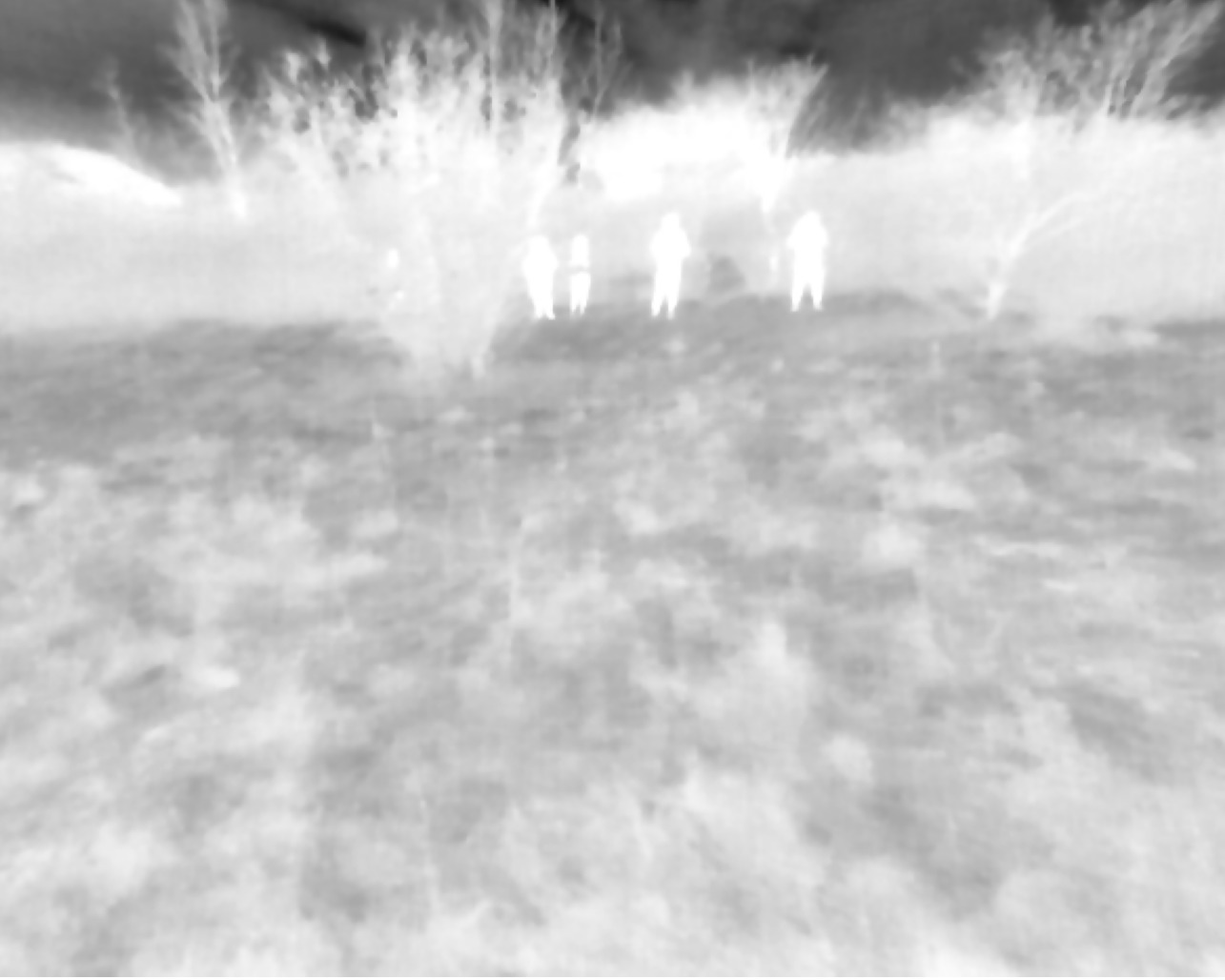}
    \end{subfigure}%
    \hspace{0.02\columnwidth}%
    \begin{subfigure}{0.31\columnwidth}
        \includegraphics[width=\linewidth]{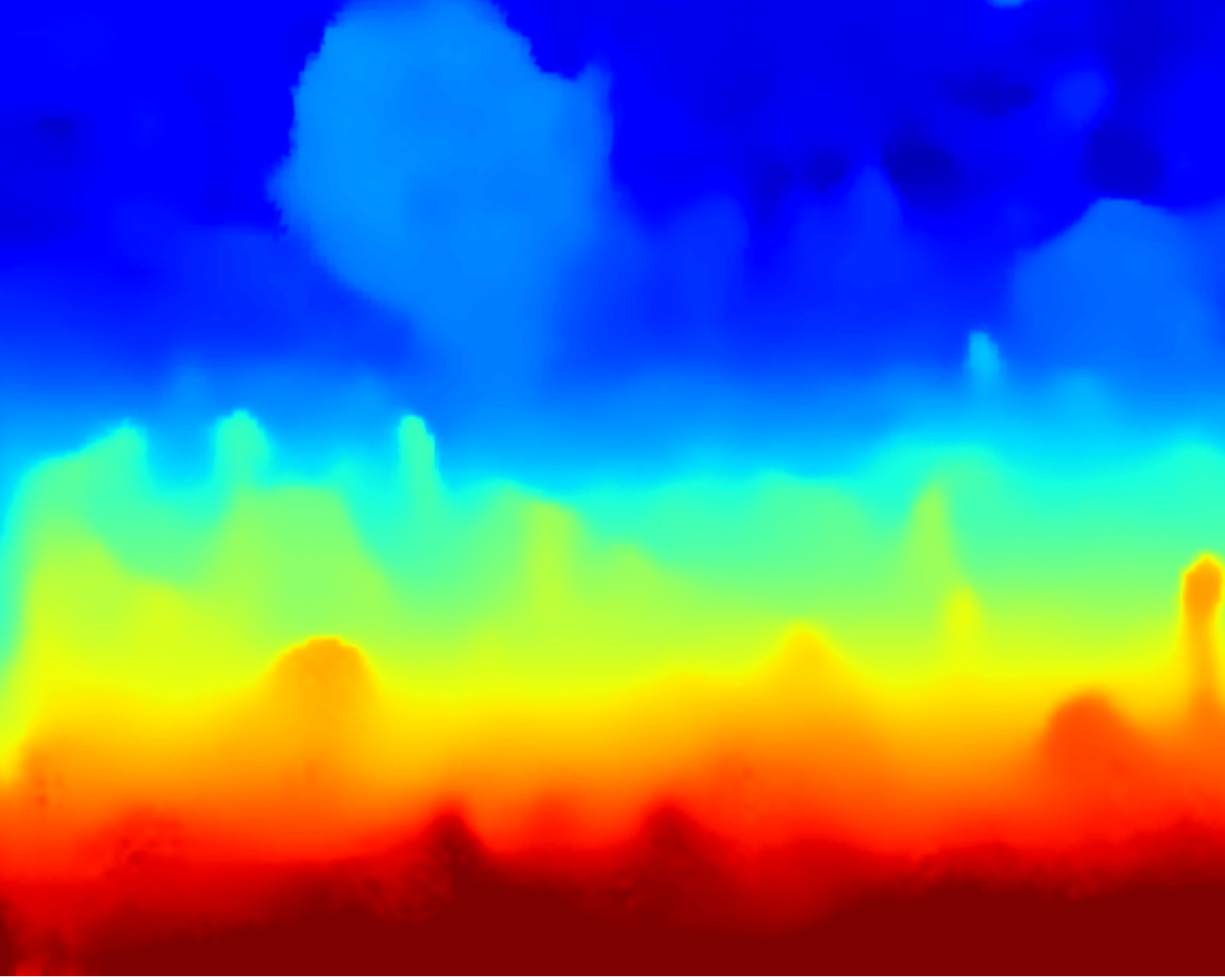}
    \end{subfigure}

    \vspace{0.4mm}

    \begin{subfigure}{0.31\columnwidth}
        \includegraphics[width=\linewidth]{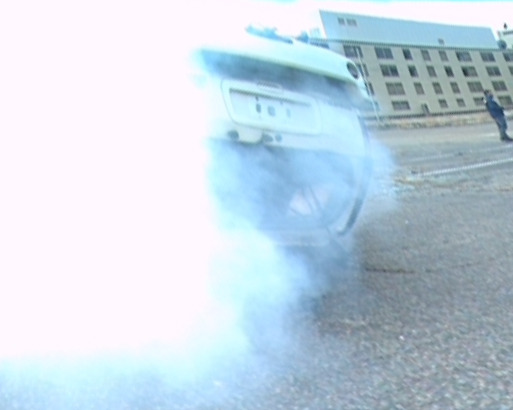}
        \caption{RGB}
    \end{subfigure}%
    \hspace{0.02\columnwidth}%
    \begin{subfigure}{0.31\columnwidth}
        \includegraphics[width=\linewidth]{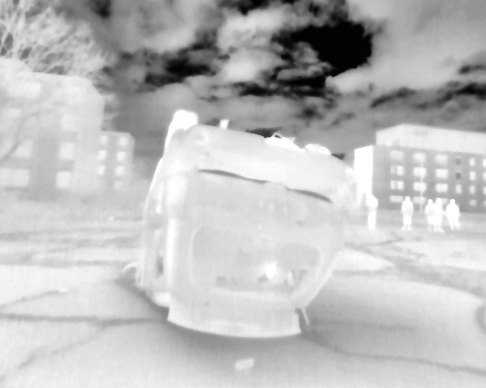}
        \caption{Thermal}
    \end{subfigure}%
    \hspace{0.02\columnwidth}%
    \begin{subfigure}{0.31\columnwidth}
        \includegraphics[width=\linewidth]{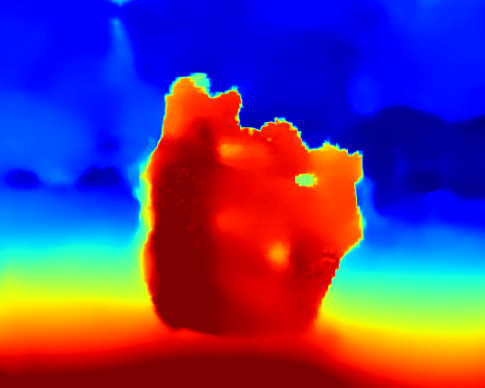}
        \caption{Estimated}
    \end{subfigure}
    \end{minipage}
    
    \caption{Qualitative inference result on unseen domain in smoke and night environments. Note the RGB sensor yields no information in the night environment, making the image black.}
    \vspace{-4pt}
    \label{fig:qualitative_smoke}
\end{figure}

%% file: text/07-conclusion.tex
\section{CONCLUSION}




In this study, we introduced the \emph{FIReStereo} dataset, specifically designed for enhancing geometric perception in visually degraded environments. This dataset leverages the capabilities of thermal long-wave infrared (LWIR) sensors, which are particularly effective in penetrating obscurities caused by darkness and smoke. Our evaluations demonstrate that models trained using the \emph{FIReStereo} dataset not only closely adhere to established depth estimation benchmarks but also show substantial improvement in handling complex environments. Notably, our dataset enables depth prediction in scenarios where traditional ground truth collection is hindered by LiDAR interference from smoke.

We hope that our work will be used in advancing robotic perception capabilities for disaster response, enabling exploration and operation in areas previously deemed inaccessible.

%% file: text/08-acknowledgement.tex
\section*{ACKNOWLEDGMENT}

This work was funded by the US Department of Agriculture under award 20236702139073. This work used Bridges-2 at PSC through allocation cis220039p from the Advanced Cyberinfrastructure Coordination Ecosystem: Services \& Support (ACCESS) program which is supported by NSF grants \#2138259, \#2138286, \#2138307, \#2137603, and \#213296.
Special thanks and appreciation are given to Shashwat Chawla, Gangadhar Nageswar, Kavin Ravie, Ishir Roongta, and Jaskaran Sodhi
for their contribution to platform design and data collection.